\documentclass[10pt, conference, compsocconf]{IEEEtran}

\usepackage{framed,multirow}
\usepackage{amssymb}
\usepackage{latexsym}
\usepackage{lipsum}
\usepackage{url,cite}
\usepackage[table]{xcolor}
\usepackage{graphicx}
\usepackage[tight,footnotesize]{subfigure}
\usepackage{dsfont}
\usepackage{color,xcolor}
\usepackage{tabularx, tabulary, multirow, bigstrut}
\usepackage[american]{babel}
\usepackage[cmex10]{amsmath}
\usepackage[vlined]{algorithm2e} %lined,boxed,commentsnumbered,boxruled
\usepackage{fixltx2e}
\usepackage[pagebackref=true,breaklinks=true,letterpaper=true,colorlinks,bookmarks=false]{hyperref}
\graphicspath{ {./figs/} }

\begin{document}

\title{Efficient Version-Space Reduction for Visual Tracking}

\author{\IEEEauthorblockN{Kourosh Meshgi, Shigeyuki Oba, Shin Ishii}
\IEEEauthorblockA{Graduate School of Informatics, Kyoto University, 
Kyoto, Japan\\
\{meshgi-k, oba, ishii\}@sys.i.kyoto-u.ac.jp}
}

\maketitle

\begin{abstract}
Discrminative trackers, employ a classification approach to separate the target from its background. To cope with variations of the target shape and appearance, the classifier is updated online with different samples of the target and the background. Sample selection, labeling and updating the classifier is prone to various sources of errors that drift the tracker. We introduce the use of an efficient version space shrinking strategy to reduce the labeling errors and enhance its sampling strategy by measuring the uncertainty of the tracker about the samples. The proposed tracker, utilize an ensemble of classifiers that represents different hypotheses about the target, diversify them using boosting to provide a larger and more consistent coverage of the version-space and tune the classifiers' weights in voting. The proposed system adjusts the model update rate by promoting the co-training of the short-memory ensemble with a long-memory oracle. The proposed tracker outperformed state-of-the-art trackers on different sequences bearing various tracking challenges.
\end{abstract}

\IEEEpeerreviewmaketitle

\section{Introduction}
\label{sec1}

Visual tracking is one of the fundamental problems in computer vision that has many applications ranging from action recognition and human-computer interfaces to autonomous navigation and robotics. The most general type of tracking is single-object model-free online tracking, in which the object is annotated in the first frame, and tracked in the subsequent frames with no prior knowledge about the target's appearance, its motions, the background, the configurations of the camera, and other conditions of the scene. Visual tracking is still considered as a challenging problem despite that many efforts have been made to address abrupt appearance changes of the target \cite{bao2012real}, its sophisticated transformations \cite{kwon2011tracking} and deformations \cite{hare2011struck}, background clutter \cite{dinh2011context}, occlusion \cite{meshgi2016data}, and motion artifacts \cite{wu2011blurred}. 

Earlier visual tracking approaches aimed to construct a generative model for the target by using statistical models \cite{comaniciu2003kernel,perez2002color,oron2015locally}, robust features \cite{he2013visual} and resilient representations (e.g., compressive sensing \cite{zhang2012real} and sparse representations \cite{bao2012real}), and online learning of subspace models \cite{ross2008incremental} and visual dictionaries \cite{taalimi2015online}. Recent approaches reformulates the problem, as a foreground versus background classification, utilizing one or more classifiers \cite{avidan2007ensemble}, or discrminative filters \cite{henriques2012exploiting,henriques2015high,kiani2015correlation}. Among such discriminative approaches, tracking-by-detection algorithms (e.g., \cite{babenko2009visual}) have emerges, focusing on the distinguishing of the target object from its background by matching candidates against the incrementally learned model of the target.

Despite the remarkable performance of tracking-by-detection schemes in large benchmarks \cite{wu2013online,kristan2015visual}, they suffer from several shortcomings, which prevent these trackers to perform to their potential: \textit{(i)} equal weights for all training samples, while different samples have different levels of similarity to the target/background and should not be treated the same \cite{henriques2012exploiting}; \textit{(ii)} label noise problem, since the labelers are usually designed using heuristics, rather than being tightly-coupled to the classifier \cite{hare2011struck}. \textit{(iii)} self-learning loop, which means that the classifier is retrained using its own output from the earlier tracking episodes, which amplitudes a training noise in the classifier and accumulate the error over time. \textit{(iv)} stability-plasticity dilemma \cite{grabner2008semi}, that set a complex equilibrium for the classifier's rate of update, and \textit{(iv)} model update strategy \cite{matthews2004template}, that if not designed well, may lead to the loss of essential information during the model update.

\begin{figure}
\includegraphics[width=1\linewidth]{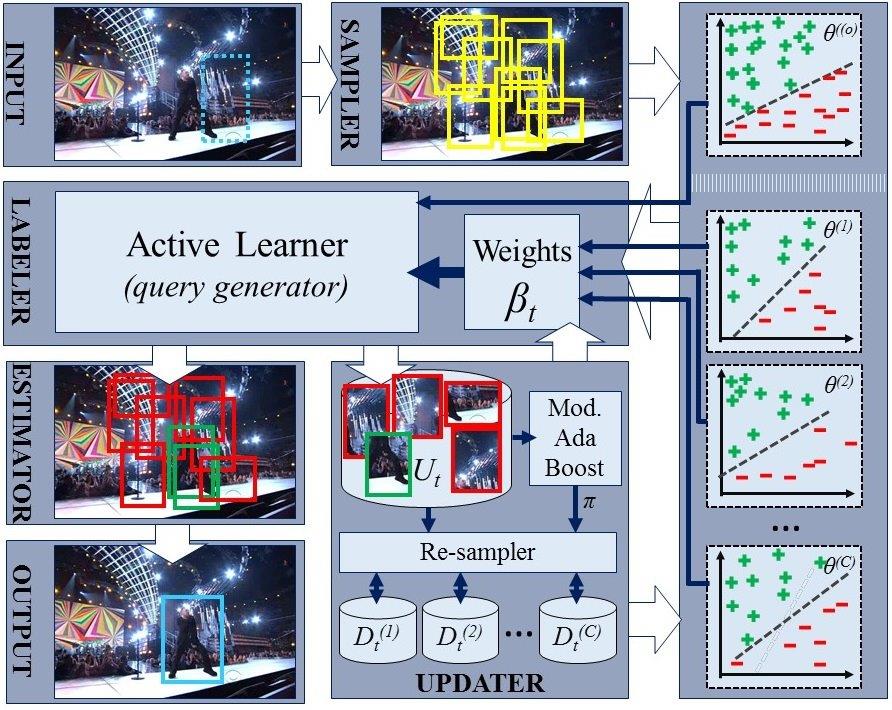}
\caption{Schematic of the system. The proposed tracker, QBST, collects samples around the last known target position and send them to the active labeler, which performs a boosting on an ensemble of randomized classifiers, and queries the disputed samples from an ``oracle''. The labels are then propagated to the next stage, where the state of the target is estimated. Finally, the ensemble classifiers of the system are updated in a query-by-boosting \cite{abe1998query} fashion, in which a modified AdaBoost adjust the weights of the classifiers, To robustify the tracker against motion and appearance jitters, the complete classifier is updated with longer intervals.}
\label{fig:schematic}
\vspace{-0.5 cm}
\end{figure}

To address these issues, various solutions have been introduced in the literature, yet a comprehensive solution is to be proposed. Adaptive weights for the samples based on their appearance similarity to the target \cite{perez2002color}, occlusion state \cite{kwak2011learning,meshgi2016data}, and spatial distance to previous target location \cite{wu2015robust} have been considered, especially in the context of tracking-by-detection, boosting \cite{freund1995desicion,oza2000online} have been extensively investigated \cite{grabner2006real,leistner2009robustness,babenko2009visual}. 
On the other hand, mistakes of the labeler manifest themselves as label noise that confuses the classifier. To tackle this problem researchers utilizes robust loss functions \cite{leistner2009robustness,masnadi2010design}, semi-supervised \cite{grabner2008semi,leistner2009semi} and multi-instance schemes \cite{babenko2009visual,zeisl2010online}, efficient sparse sampling \cite{henriques2015high}, context information \cite{grabner2010tracking,dinh2011context}, sample informativeness for the classifier \cite{zhang2013robust}, landmark-based label propagation \cite{wu2015robust}, and some of them even reformulate the tracking-by-detection pipeline to combine the labeling and learning process \cite{hare2011struck}.
The problem amplifies when the tracker does not have a forgetting mechanism or a way to obtain external scaffolds. This inspired the use of co-tracking \cite{tang2007co}, ensemble tracking \cite{saffari2009line,zhang2014meem} or label verification schemes \cite{kalal2012tracking} to break the self-learning loop using auxiliary classifiers.
Updating the classifier is another challenge of the tracking-by-detection schemes. Replacing the weakest classifier of an ensemble \cite{grabner2006real} or the oldest one \cite{avidan2007ensemble}, budgeting the sample pool of the classifier \cite{hare2011struck}, and co-learning \cite{tang2007co} were among the most popular updating methods in the literature. On top of that, the frequency of update is another important role-player in tracker's performance \cite{grabner2008semi}. Higher update rates capture the rapid target changes, but is prone to occlusions, whereas slower update paces provide a long memory for the tracker to handle temporal target variations but lack the flexibility to accommodate permanent target changes. To this end, researchers try to combine long- and short-term memories \cite{hong2015multi} or role-back improper updates \cite{zhang2014meem}.

Traditionally, ensemble trackers were used to providing a multi-view classification of the target, realized by using different features to construct weak classifiers. In this view, different classifiers represent different hypotheses in the version-space, attempting to accurately model the target appearance. Such hypotheses are highly-overlapping, therefore an ensemble of them overfits the target. A desired committee, however, consists of competing hypotheses, all consistent with the training data, but each of the specialized in certain aspects of the target. In this view, the most informative data samples are those about which the hypotheses disagree the most, and by labeling them the version-space is minimized resulting in a quick-convergence yet accurate classification \cite{seung1992query}.  Motivated by this, we proposed a tracker that employs a diverse ensemble of classifiers and selects the most informative data samples to be labeled.
%Another, more theoretically-motivated query selection framework is the query-by-committee (QBC) algorithm (Seung et al., 1992). The QBC approach involves maintaining a committee of models which are all trained on the current labeled set L, but represent competing hypotheses. Each committee member is then allowed to vote on the labelings of query candidates. The most informative query is considered to be the instance about which they most disagree. The fundamental premise behind the QBC framework is minimizing the version space, which is the set of hypotheses that are consistent with the currently labeled training data L{\color{red}\lipsum[1]}
%\begin{figure}[!t]
%\includegraphics[width=1\linewidth]{ph}
%\caption{Schematic}
%\label{fig:space}
%\vspace{-0.5 cm}
%\end{figure}

In this study, we propose the query-by-boosting tracker, QBST, a hybrid discriminative tracker, that employs the concept of co-tracking, in which a diverse short-memory ensemble of classifiers (the \textit{committee}) exchange information with a long-memory classifier (the \textit{oracle}) to reduce label-noise, balance the stability-plasticity equilibrium, and produce high-accuracy tracking results. This information exchange is built upon the concept of query-by-boosting \cite{abe1998query}, in which the optimal data to exchange is selected as the sample which the weighted majority voting by the committee has the least margin. The model update is performed by boosting each classifier to maintain a diverse committee, yet building an accurate strong classifier.

In the next section, a tracking-by-detection pipeline is formulated and extended to ensemble tracking case, to establish the notion that our proposed Query-by-Boosting Tracker. Section \ref{sec3} illustrates the experimental design and criteria. This section elaborates on the state-of-the-art trackers used in this study and presents the results of the proposed tracker that highlight its superior performance. The manuscript is concluded in Section \ref{sec:conclusion}. 

\section{Proposed Method}
\label{sec2}
In the following section, an overview of adaptive tracking-by-detection approaches is provided, and the proposed framework is elaborated.

\subsection{Tracking by Detection}
By definition, a tracker tries to determine the state of the target $\mathbf{p}_t$ in frame $F_t$ ($t \in \{1,\ldots,T\}$) by finding the transformation $\mathbf{y}_t$ from its previous state $\mathbf{p}_{t-1}$. In tracking-by-detection formulation, the tracker employs a classifier $\theta_t$ to separate the target from the background. It is realized by evaluating possible candidates from the expected target state-space $\mathcal{Y}_t$. The candidate whose appearance resembles the target the most, is usually considered as the new target state. More specifically, the candidate that maximizes the classification score, is believed to be the new target. Finally, the classifier is updated to reflect the recent changes in the target as well as the background.

To this end, first several samples $\mathbf{x}_t^{\mathbf{p}_{t-1} \circ \mathbf{y}_t^j}  \in \mathcal{X}_t$ are obtained by a transformation $\mathbf{y}^j_t \in \mathcal{Y}_t$ from the previous target state, $\mathbf{p}_{t-1} \circ \mathbf{y}^j_t$. Sample $j$ indicates the location $\mathbf{p}_{t-1} \circ \mathbf{y}_t^j$ in the frame $F_t$, where the image patch $\mathbf{x}_t^{\mathbf{p}_{t-1} \circ \mathbf{y}_t^j}$ is contained. The transformation space $\mathcal{Y}_t$ is usually defined by motion models, optical flow, context supports \cite{grabner2010tracking}, confidence maps \cite{tang2007co}, or a combination of these. Then, each sample is evaluated by the classifier scoring function $h: \mathcal{X}_t \rightarrow \mathbb{R}$.
\begin{equation}
\mathbf{s}^j_t = h(\mathbf{x}_t^{\mathbf{p}_{t-1} \circ \mathbf{y}_t^j} | \theta_t).
\label{eq:score_single}
\end{equation}

This score is utilized to obtain a label $\ell^j_t$ for the sample, typically by thresholding its score,
\begin{align}
\ell^j_t &=
  \begin{cases}
   +1        & ,\mathbf{s}^j_t > \tau_u \\
   -1        & ,\mathbf{s}^j_t < \tau_l \\
   0         & ,\text{otherwise}
  \end{cases}
  \label{eq:label_single}
\end{align}
where $\tau_l$ and $\tau_u$ serves as lower and upper bounds respectively. In supervised learning schemes (e.g., \cite{grabner2006real}), these thresholds are equal ($\tau_l = \tau_u$), whereas by employing semi-supervised learning for the classifier (e.g., \cite{grabner2008semi,saffari2010robust}), the trackers allow for some samples to be unlabeled. In addition, trackers based on multi-instance learning (e.g., \cite{babenko2009visual,zhang2013robust}), bag the samples and apply a label on each bag to handle ambiguity of labeling, and active-learning trackers (e.g., \cite{meshgi2016robust}) rely on their oracle for disambiguation. 

Finally, the target location $\mathbf{y}_t$ is obtained by comparing the samples classification scores. To obtain the exact target state, the sample with highest score is selected as the new target, yet it is only realistic if a dense sampling is employed. 
\begin{align}
\mathbf{y}_t = \underset{\mathbf{y}_t^j \in \mathcal{Y}}{\mathrm{argmax}} \; ( s_t^j ) = \underset{\mathbf{y}_t^j \in \mathcal{Y}}{\mathrm{argmax}} \; \big( h(\mathbf{x}_t^{\mathbf{p}_{t-1} \circ \mathbf{y}_t^j} | \theta_t) \big),
\label{eq:localize_single}
\end{align}
If the number of samples are limited ($\{ \mathbf{y}^1_t, \ldots, \mathbf{y}^n_t \}$), the approximate target location $\mathbf{\hat{y}}_t$ is obtained by maximizing the expectation of target, i.e., by taking a weighted average of the target candidates (i.e., positive samples).
\begin{align}
\mathbf{\hat{y}}_t = \;\mathbb{E} [ \mathbf{y}_t^j ] = \sum_{\forall j, \ell^j_t>0} \mathbf{s}^j_t \mathbf{y}_t^j.
\label{eq:localize_approx}
\end{align}

A subset of the samples and their labels are used to re-train the classifier's model $\theta_t$,
\begin{equation}
\theta_{t+1} = u(\theta_t, \mathcal{X}_{\xi(t)}, \mathcal{L}_{\xi(t)})
\end{equation}
in which $\mathcal{L}_t$ denotes the set of labels of the samples $\mathcal{X}_t$, $u(.)$ is the model update function, and the $\xi(t)$ defines the subset of the samples that the tracker considers for updating its model. Many of the adaptive trackers utilize online-learning classifiers \cite{hare2011struck,ross2008incremental} in which only the data from the recent frame ($\xi(t)=\{t\}$) is used. Fixed trackers use only the data from the first frame ($\xi(t) = {1}$) and some trackers utilize the samples obtained from several recent frames to update their model ($\xi(t) = \{t-\Delta, \ldots, t-1, t \}$).

\subsection{Ensemble Discriminative Tracking}
An ensemble discriminative tracker employs a set of classifiers instead of one. These classifiers, hereafter called \textit{committee}, are represented by $\mathcal{C}=\{\theta_t^{(1)},\ldots,\theta_t^{(C)}\}$, and are typically homogeneous and independent (e.g.,  \cite{saffari2009line,leistner2010miforests}). Popular ensemble trackers utilize the majority voting of the committee as their utility function,
\begin{equation}
\mathbf{s}^j_t = \sum_{c=1}^C \mathrm{sign} \big( h(\mathbf{x}_t^{\mathbf{p}_{t-1} \circ \mathbf{y}_t^j} | \theta_t^{(c)}) \big).
\label{eq:score_ensemble}
\end{equation}
and eq\eqref{eq:label_single} is used to label the samples. Another approach is to use boosting to construct a robust classifier from several weak committee members (e.g., \cite{grabner2006real}),
\begin{equation}
\mathbf{s}^j_t = \sum_{c=1}^C \alpha_t^{(c)} \mathrm{sign} \big( h(\mathbf{x}_t^{\mathbf{p}_{t-1} \circ \mathbf{y}_t^j} | \theta_t^{(c)}) \big),
\label{eq:score_ensemble_boost}
\end{equation}
where $\alpha_t^{(c)}$ is the weight of classifier $c$ in this linear combination.
Finally, the model is updated for each classifier independently,
\begin{equation}
\theta^{(c)}_{t+1} = u(\theta^{(c)}_t, \mathcal{X}_{\xi(t)}, \mathcal{L}_{\xi(t)})
\label{eq:update_ensemble}
\end{equation}
meaning that all of the committee members are trained with a similar set of samples and a common label for them.
On the other hand, in co-tracking algorithms (such as \cite{tang2007co}), different classifiers have different sample set $\mathcal{X}^{(c)}_{\xi(t)}$ and label them based on their own models ($\mathcal{L}^{(c)}_{\xi(t)}$). 

\subsection{Query-by-Boosting Tracker (QBST)}
\label{sec:qbst}
The proposed tracker, QBST, is consisted of an ensemble of classifiers $\mathcal{C}$ that query the label of its most uncertain samples from another classifier, the oracle. The samples are obtained by adding a Gaussian motion model to the target last state. The weights of the committee, are adjusted regarding their labeling accuracy in classifying already labeled samples, as will discussed later. For now, let's assume the weight of committee member $c$ is equal to $\mathrm{log}(\frac{1}{\beta_t^{(c)}})$. So the committee score will be calculated as
\begin{equation}
\mathbf{s}^j_t = \sum_{c=1}^C \mathrm{log}(\frac{1}{\beta_t^{(c)}}) \; \mathrm{sign} \big( h(\mathbf{x}_t^{\mathbf{p}_{t-1} \circ \mathbf{y}_t^j} | \theta_t^{(c)}) \big).
\label{eq:score_qbst}
\end{equation}

If the committee comes to a solid vote about a sample ($\mathbf{s}^j_t > \tau_u$ or $\mathbf{s}^j_t < \tau_l$), then the sample is labeled accordingly. However, when the committee disagrees about a sample (i.e., mostly casting contradictory votes, e.g., only 4 out of 7 believes the samples contains the target), its label is queried from the oracle and the sample is added to the uncertain samples list $\mathcal{U}_t$:
\begin{align}
\ell(\mathbf{s}^j_t) &=
  \begin{cases}
   +1                                                                        & ,\mathbf{s}^j_t > \tau_u \\
   -1                                                                        & ,\mathbf{s}^j_t < \tau_l \\
  \mathrm{sign} \big( h(\mathbf{x}_t^{\mathbf{p}_{t-1} \circ \mathbf{y}_t^j} | \theta^{(o)}_t) \big)    & ,\text{otherwise}
  \end{cases}
  \label{eq:label_qbst}
\end{align}
and $\mathcal{U}_t = \{\langle \mathbf{x}^{\mathbf{p}_{t-1} \circ \mathbf{y}^j_t} , \ell(\mathbf{s}^j_t) \rangle | \tau_l \leq s_t^j \leq \tau_u \}$. Setting $\tau_l = \tau_u$ in this co-tracking framework simplifies it back to the case of a single-classifier tracking-by-detection and removes the role of the oracle.
Having the label of all samples, the target's state is estimated using eq\eqref{eq:localize_approx}.

To make the committee, in $t=1$ that the target is annotated by the user, we obtain $m'$ samples, positives (by perturbing the target with a Gaussian noise in location and scale up to 5 pixels) and negatives (the local and global background). For each committee member we select $m$ of these samples by uniform distribution ($ m \ll m'$), and form the classifier $\theta_1^{(c)}$. The oracle $\theta^{(o)}$, is initialized with all of these $m'$ samples. Later, in time $t$ the model of committee member $c$, $\theta_t^{(c)}$, is updated using a subset of $m$ samples from $\mathcal{U}_t$ such that the updated model $\theta_{t+1}^{(c)}$ have the maximum accuracy on all of the samples represented by committee member $c$. This is realized by using a modified version of AdaBoost detailed in Algorithm \eqref{alg:boost}.
\begin{equation}
\theta^{(c)}_{t+1} = u(\theta^{(c)}_t, \mathcal{X}^{(c)}_{1..t} \cup \mathcal{U}_t, \mathcal{L}^{(c)}_{1..t})
\end{equation}
As a byproduct of this boosting process, the value of $\beta_{t+1}^{(c)}$ is obtained that is used to calculate the weight of committee member $c$ in eq \eqref{eq:score_qbst}. Note that for certain samples (those not in $\mathcal{U}_t$), the committee was unanimous about the label and adding them to the training set of the committee classifiers doesn't add to the classification accuracy, while limiting the generalization. It is prudent to note that, labeling errors by the oracle, even if happens, does not distract the committee since the partial re-sampling and classifier re-weighting mechanisms undermines it significantly.
Finally, to maintain a long-term memory and slower update rate for oracle, it is updated every $\Delta$ frames with all of the samples (not only the uncertain ones).
\begin{align}
\theta^{(o)}_{t+1} = 
    \begin{cases}
    u(\theta^{(o)}_t, \mathcal{X}_{1..t}, \mathcal{L}_{1..t})     &, \mathrm{if} \; t \neq k\Delta \\
    \theta^{(o)}_{t}                                             &, \mathrm{if} \; t = k\Delta
    \end{cases}
\label{eq:update_qbst}
\end{align}

Algorithm \eqref{alg:qbst} summarizes the proposed tracker.

%%%%%%%%%%%%%%
\begin{algorithm}[!t]
\DontPrintSemicolon
\SetKwInOut{Input}{input}\SetKwInOut{Output}{output}
\Input{Committee weights $\beta_{t}^{(c)}$}
\Input{Committee models $\theta_{t}^{(c)}$, Oracle model $\theta^{(o)}$}
\Input{Committee samples $\mathcal{D}_{t}^{(c)}$, Oracle samples $\mathcal{D}^{(o)}$}
\Input{Target position in previous frame $\mathbf{p}_{t-1}$}
\Output{Target position in current frame $\mathbf{p}_{t}$}
\BlankLine
\For{$j \leftarrow 1$ \KwTo $n$ }
{
\emph{Sample a transformation $\mathbf{y}_t^j \sim \mathcal{N}(\mathbf{p}_{t-1},\Sigma_{search})$}\;
\emph{Calculate QBST committee score $s_t^j$} (eq\eqref{eq:score_qbst})\;
\uIf(sample label is uncertain){$\tau_l \leq s_t^j \leq \tau_u$}
        {
        \emph{$\ell_t^j = \mathrm{sign} \big( h(\mathbf{x}_t^{\mathbf{p}_{t-1} \circ \mathbf{y}_t^j} | \theta^{(o)}) \big)$}\;
        \emph{$\mathcal{U}_t \leftarrow \mathcal{U}_t \cup \{\langle \mathbf{x}^{\mathbf{p}_{t-1} \circ \mathbf{y}^j_t} , \ell^j_t \rangle\}$}\;
        }
\Else
        {
        \emph{$\ell_t^j = \mathrm{sign} (s_t^j)$}\;
        }
\emph{$\mathcal{D}^{(o)} \leftarrow \mathcal{D}^{(o)} \cup \{\langle \mathbf{x}^{\mathbf{p}_{t-1} \circ \mathbf{y}^j_t} , \ell^j_t \rangle\}$}\;
}% endloop sample 
\For{$c \leftarrow 1$ \KwTo $C$ }
{
\emph{ $\mathcal{D}_{t+1}^{(c)},\beta_{t+1}^{(c)} \leftarrow ModAdaBoost (\mathcal{U}_t,\mathcal{D}_{t}^{(c)},m)$ }(alg\ref{alg:boost})\;
\emph{ $\theta_{t+1}^{(c)} \leftarrow u(\theta_t^{(c)}|\mathcal{D}_{t+1}^{(c)})$}\;
}
\If{$\mathrm{mod}(t,\Delta^{(o)}) = 0$}
{
\emph{Retrain $\theta^{(o)}$ with $\mathcal{D}^{(o)}$}\;
}% end if
\emph{Estimate transformation $\mathbf{\hat{y}}_t$} (eq\eqref{eq:localize_approx}) \;
\emph{Calculate target position $\mathbf{p}_t = \mathbf{p}_{t-1} \circ \mathbf{\hat{y}}_t$}\;
\BlankLine
\caption{Query-by-Boosting Tracker (QBST)}
\label{alg:qbst}
\vspace{-0.1em}
\end{algorithm}
%%%%%%%%%%%%%%
%
%%%%%%%%%%%%%%
\newcommand\mycommfont[1]{\footnotesize\ttfamily\textcolor{black}{#1}}
\SetCommentSty{mycommfont}

\begin{algorithm}[!t]
\DontPrintSemicolon
\SetKwInOut{Input}{input}\SetKwInOut{Output}{output}
\Input{Samples to be added to each committee $m$}
\Input{Uncertain Samples $\mathcal{U}_t = \{(\mathbf{x}_1,\ell_1),...,(\mathbf{x}_n,\ell_n) \}$}
\Input{Committee samples $\mathcal{D}_{t}^{(c)}$}
\Output{Extended Committee samples $\mathcal{D}_{t+1}^{(c)}$}
\Output{Committee weights $\beta_{t+1}^{(c)}$}
\BlankLine
\emph{$\pi_1(\mathbf{x}_i) = \frac{1}{n+m} , \,\,\,  i \in \{1,\ldots,n+m\}$}\;
\For{$\tau \leftarrow 1$ \KwTo $\tau_B$ }
{
\emph{$S_\tau \leftarrow m$ samples from $\mathcal{U}_t$ w.r.t. $\pi_\tau$  }\;
\emph{Train model $\theta'_\tau$ on $S_\tau \cup \mathcal{D}_{t}^{(c)}$}\;
\emph{Compute error rate $\epsilon_\tau = \sum_{h(\mathbf{x}_i|\theta'_\tau) \neq \ell_i} \pi_\tau(\mathbf{x}_i)$}\;
\emph{Calculate $\beta_\tau = \frac{\epsilon_\tau}{1- \epsilon_\tau}$}\;
\emph{Update the re-sampling distribution $\pi_{\tau+1}$}:
\begin{align}
\pi_{\tau+1}(\mathbf{x}_i) = 
\begin{cases}
  \frac{1}{Z}\pi_{\tau}(\mathbf{x}_i)\beta_\tau     & ,\mathrm{if} \; h(\mathbf{x}_i|\theta'_\tau) = \ell_i  \nonumber  \\
  \pi_{\tau}(\mathbf{x}_i)                             & ,\text{otherwise} \nonumber \\
\end{cases}
\end{align}
\tcc{Constant $Z$ satisfies $\sum_i \pi_{\tau+1}(\mathbf{x}_i) = 1$}
}
\emph{$\tau^* \leftarrow \mathrm{argmin}_\tau \; \epsilon_\tau$}\;
\emph{$\beta_{t+1}^{(c)} \leftarrow \beta_{\tau^*}$}\;
\emph{$\mathcal{D}_{t+1}^{(c)} \leftarrow \mathcal{D}_{t}^{(c)} \cup S_{\tau^*}$}\;

\BlankLine
\caption{Modified AdaBoost Algorithm}
\label{alg:boost}
\end{algorithm}
%%%%%%%%%%%%%%

\subsection{Considerations}
\label{sec:consider}
There are several parameters in the system such as the number of committee members ($C$), parameters of sampling step (number of samples $n$, effective search radius $\Sigma_{search}$), boosting steps ($\tau_B$), and the holding time of oracle ($\Delta$). Except for $C$ that have been adjusted to control the speed of the tracker, the rest of parameters have been adjusted using cross-validation. Another parameter, the size of new samples to be added to each committee member, $m$, proved to be a sensitive parameter. Larger values of $m$ results in less diverse committee that reduces the role of oracle in the designed system, whereas smaller values of $m$ tend to miss the latest changes of the quick-changing target. This parameter was tuned using a simulated annealing optimization on a cross validation set (starting with $m=n/c$ as a rule of thumb). In our implementation, we used a lazy classifier and reused the calculations with a caching mechanism to accelerate the AdaBoost process and committee update. 

%%%%%%%%
\begin{figure}[!b]   
\centering
\subfigure{\includegraphics[width= 0.9\linewidth]{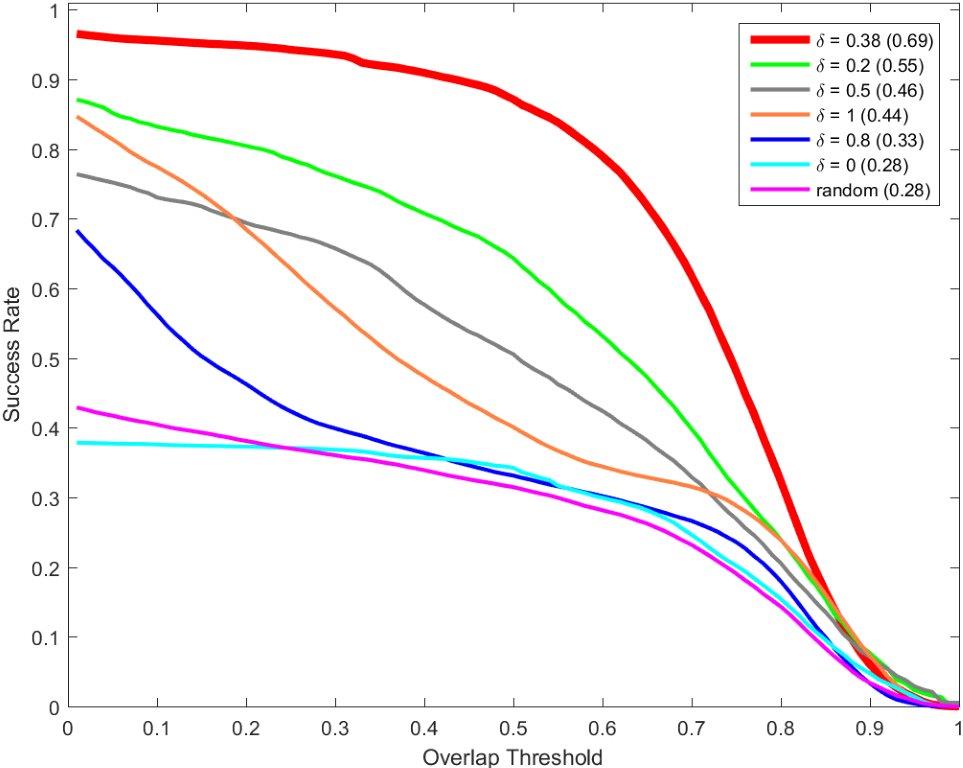}}
\caption{The effect of $\delta$ on the performance of QBST measured by $AUC$ of precision plot for all sequences in OTB-50 \cite{wu2013online}(r.t. text for discussion).}
\label{fig:delta}
\vspace{-0.5 cm}
\end{figure}
%%%%%%%%

Arguably, the most important parameters of the system are labeling thresholds ($\tau_l$ and $\tau_u$), since they control the ``activeness'' of the data exchange between the committee and the oracle. Assume that $\tau_l = -\delta$ and $\tau_l = +\delta$ ($\delta \in [0,1]$); if $\delta \rightarrow 1$ then the tracker would be a conventional binary classifier modeled by the oracle, and if $\delta \rightarrow 0$, the tracker would be governed solely based on the decision of the committee. The information exchange in one way is in the form of querying the most informative labels from the oracle, and on the other way is re-training the oracle with the labeled samples by the committee (for certain samples). We observed that this exchange is essential to construct a robust and accurate tracker. Moreover, such data exchange not only breaks the self-learning loop but also manages the plasticity-stability equilibrium of the tracker. In this view, lower values of $\delta$ correspond to a more-flexible tracker, while higher values make it more conservative.
%%%%%%%%
\begin{figure}[!t]   
\centering
\includegraphics[width= 0.48\linewidth]{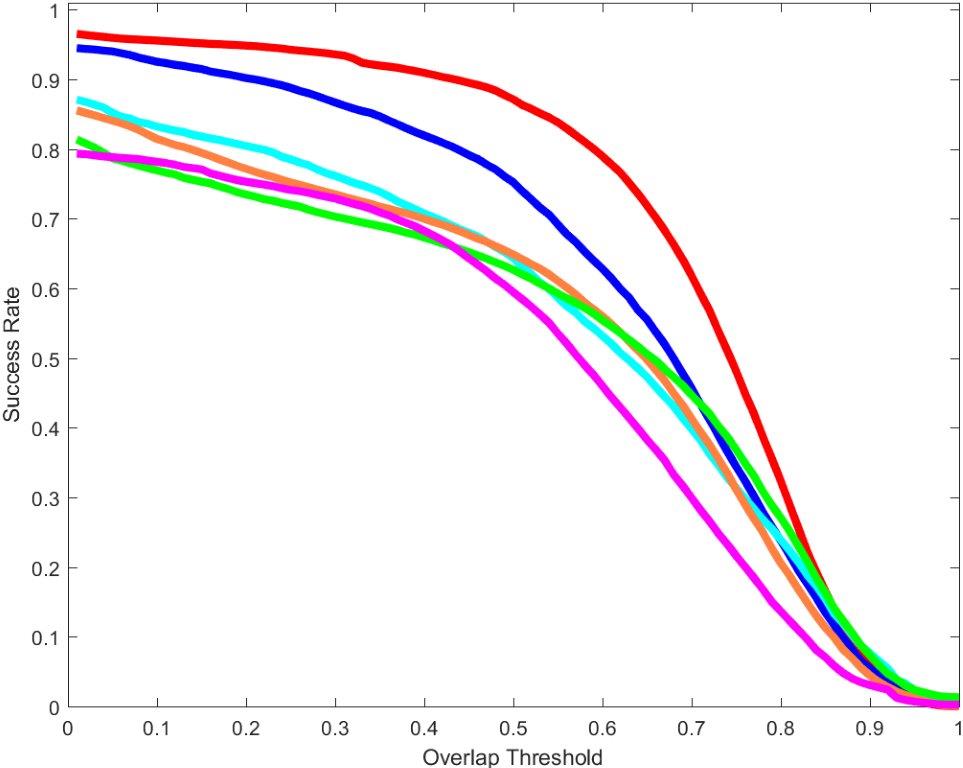}
\includegraphics[width= 0.48\linewidth]{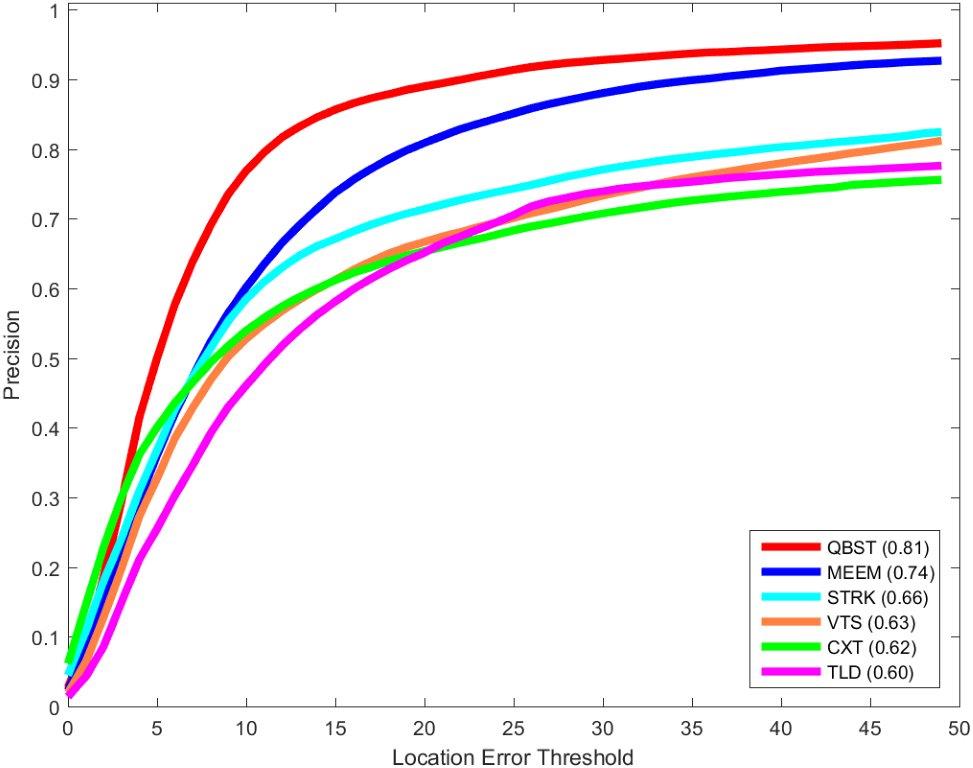}
\caption{Quantitative evaluation of trackers using success plot and precision plot for all sequences in OTB-50 \cite{wu2013online}.}
\label{fig:precision}
\vspace{-0.5 cm}
\end{figure}
%%%%%%%%

\section{Evaluation}
\label{sec3}
%%%%%%
\begin{figure*}
\centering
\subfigure[ALL (QBST, MEEM, STRK)\label{fig:all}]{\includegraphics[width= 0.24\linewidth]{success_plot_ALL.jpg}}
\subfigure[IV (QBST, MEEM, VTS)\label{fig:iv}]{\includegraphics[width= 0.24\linewidth]{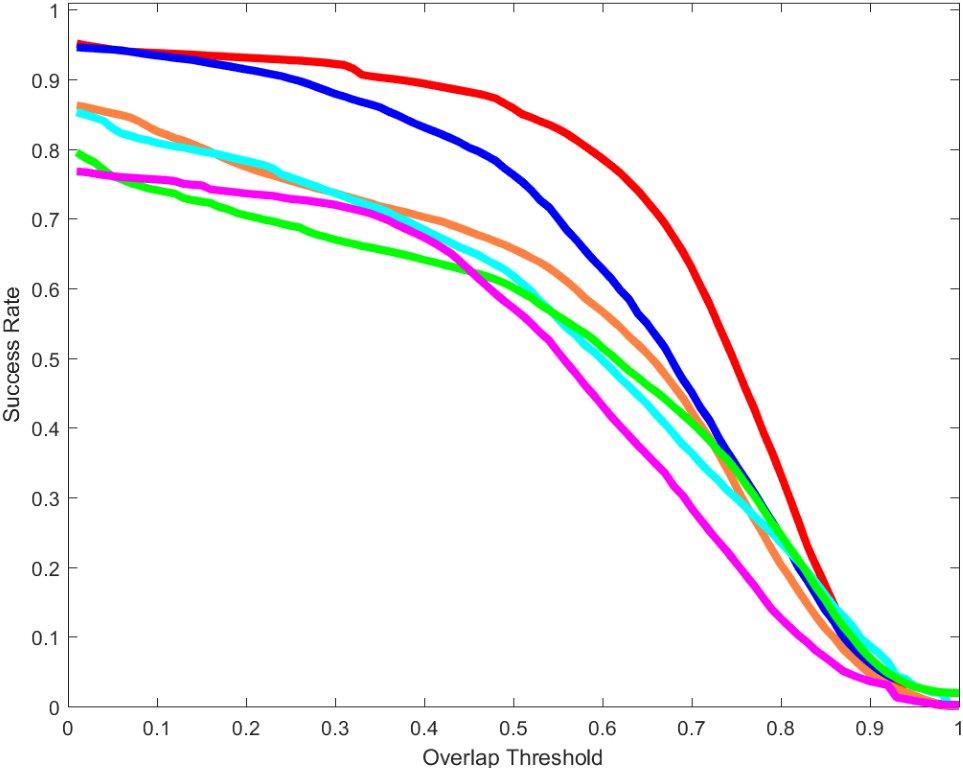}}
\subfigure[SV (QBST, MEEM, VTS)\label{fig:sv}]{\includegraphics[width= 0.24\linewidth]{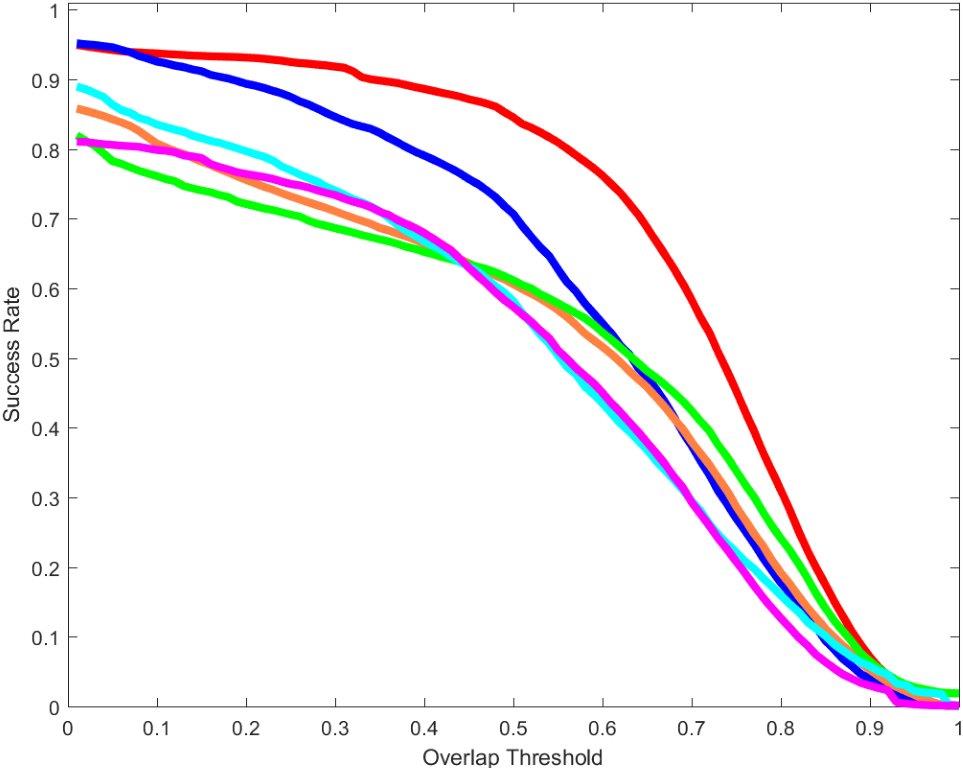}}
\subfigure[OCC (QBST, MEEM, VTS)\label{fig:occ}]{\includegraphics[width= 0.24\linewidth]{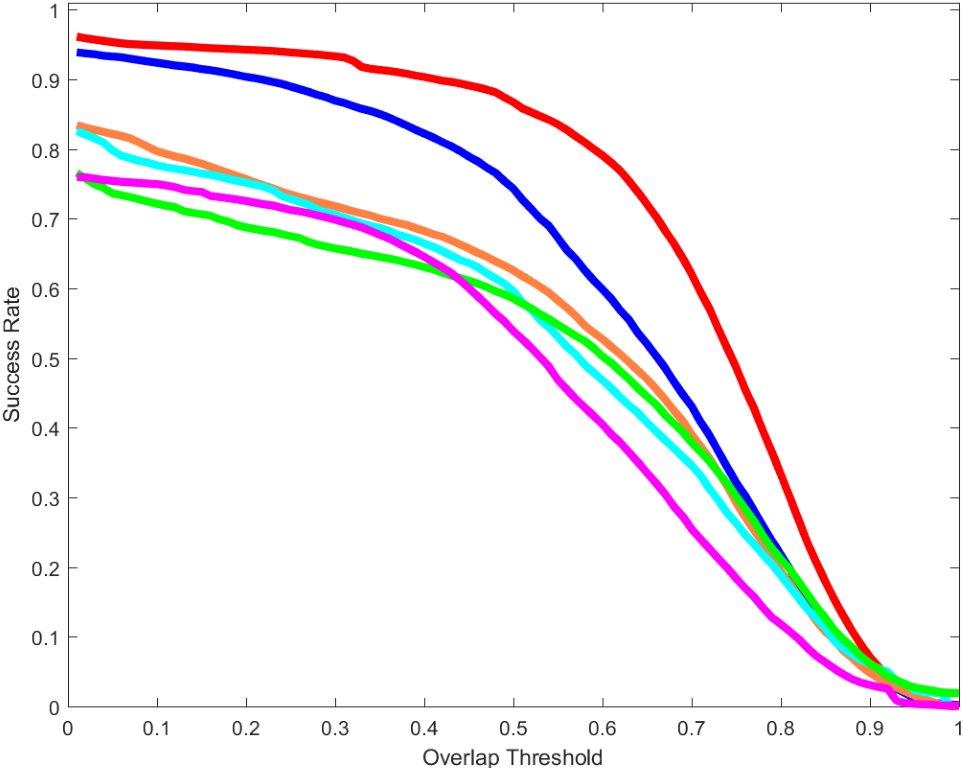}}
\subfigure[DEF (QBST, MEEM, VTS)\label{fig:def}]{\includegraphics[width= 0.24\linewidth]{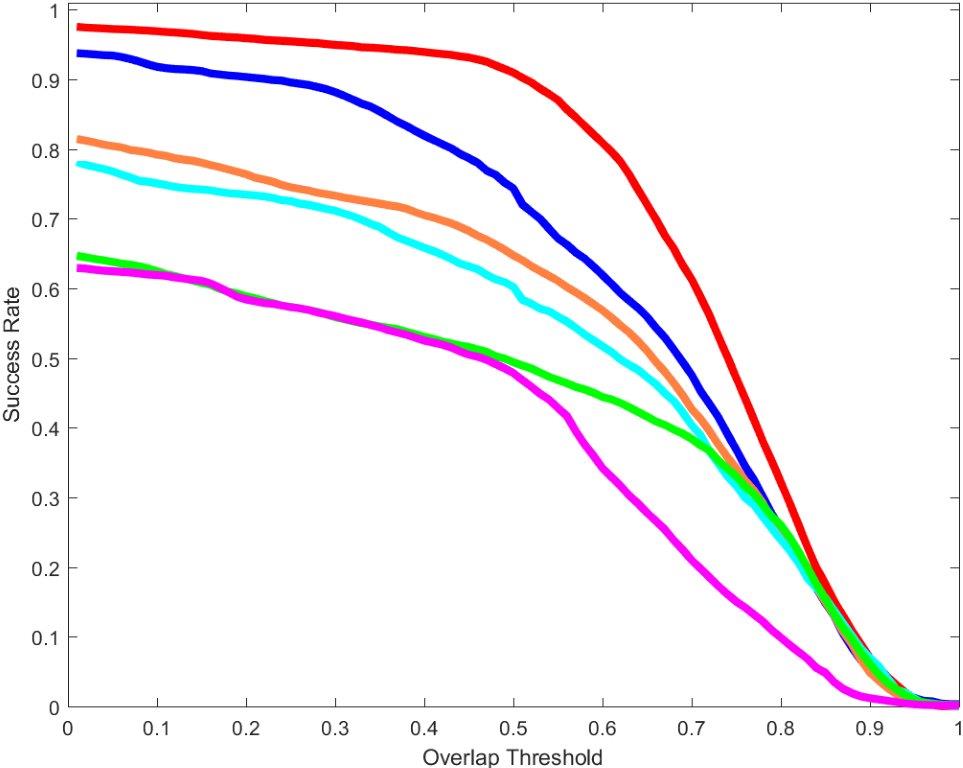}}
\subfigure[IPR (QBST, CXT, MEEM)\label{fig:ipr}]{\includegraphics[width= 0.24\linewidth]{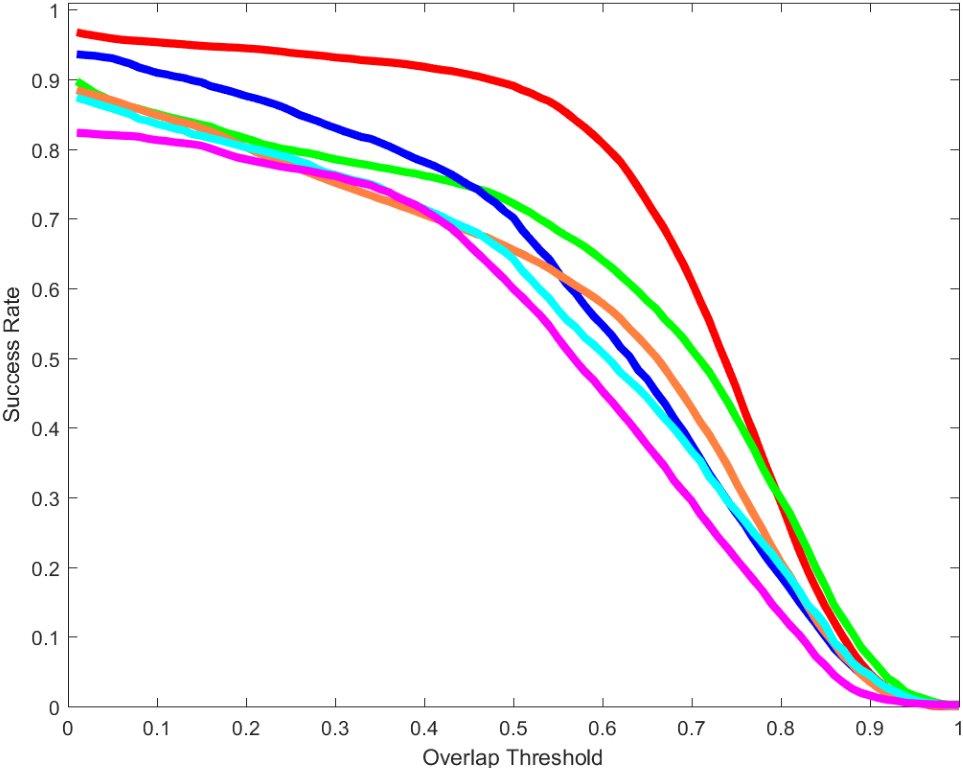}}
\subfigure[OPR (QBST, MEEM, VTS)\label{fig:opr}]{\includegraphics[width= 0.24\linewidth]{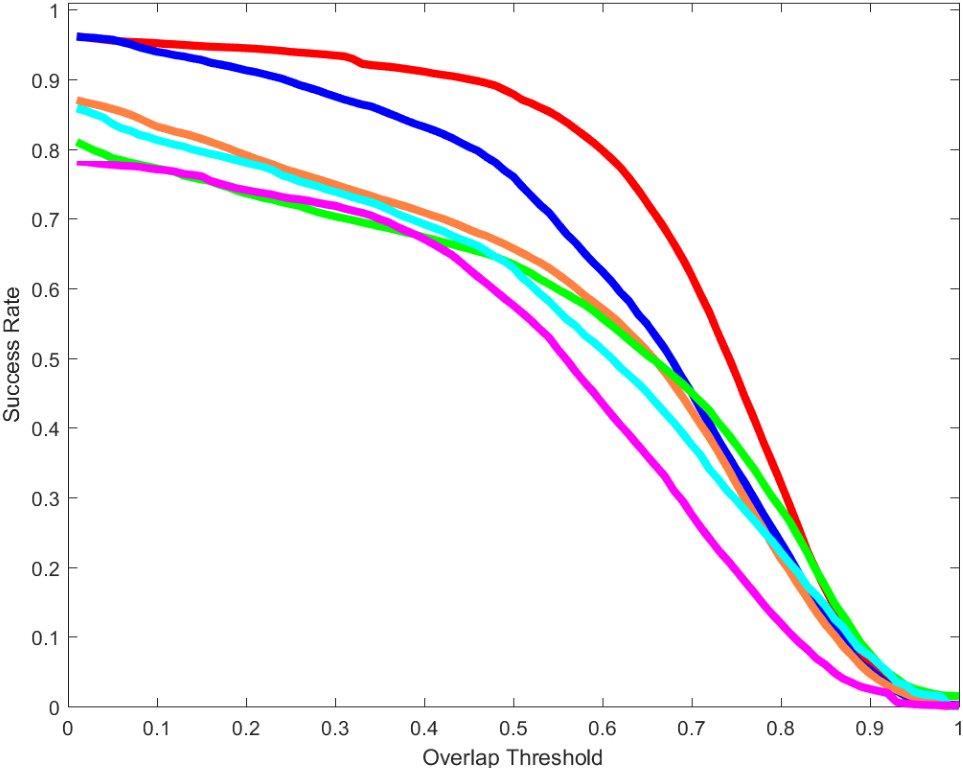}}
\subfigure[OV (QBST, MEEM, TLD)\label{fig:ov}]{\includegraphics[width= 0.24\linewidth]{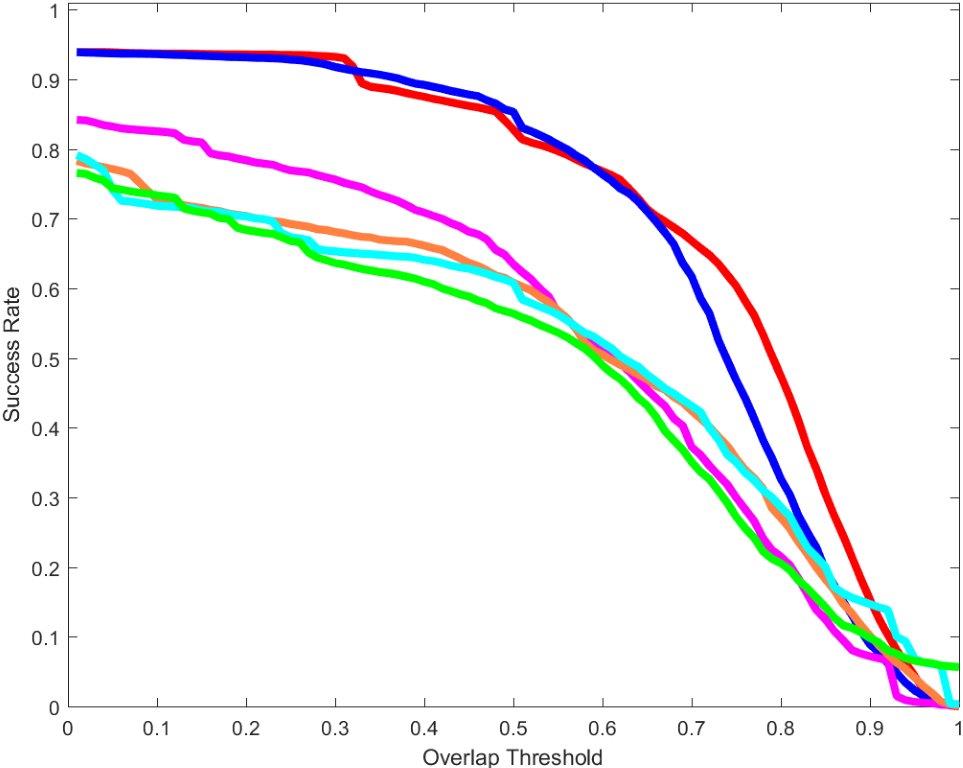}}
\subfigure[BC (QBST, MEEM, VTS)\label{fig:bc}]{\includegraphics[width= 0.24\linewidth]{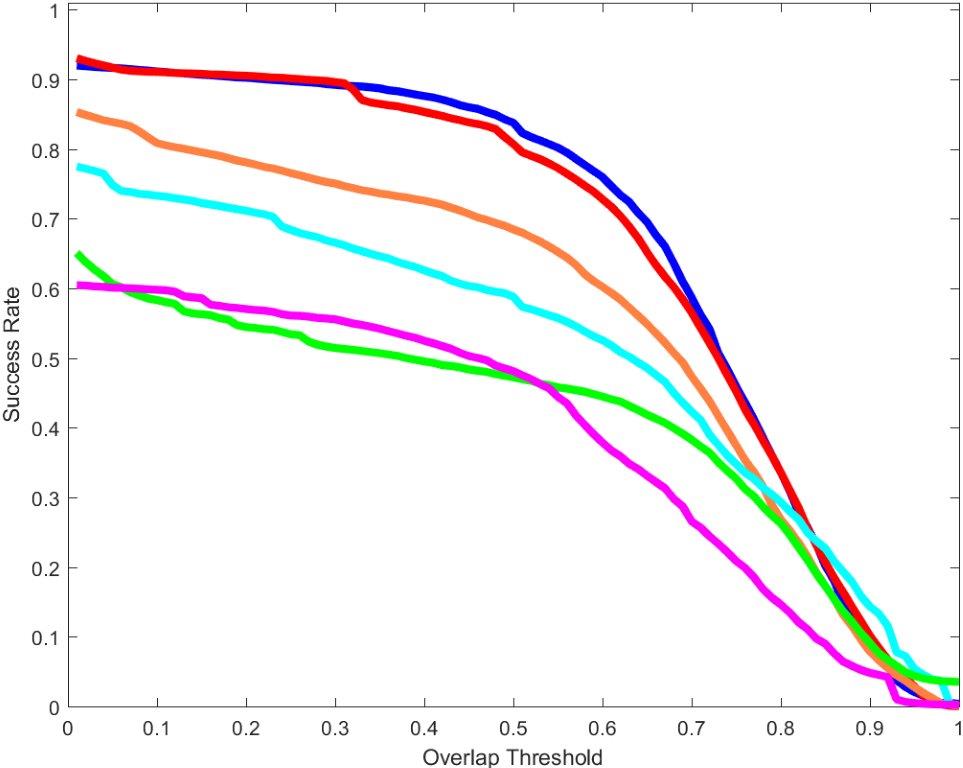}}
\subfigure[LR (QBST, MEEM, CXT)\label{fig:lr}]{\includegraphics[width= 0.24\linewidth]{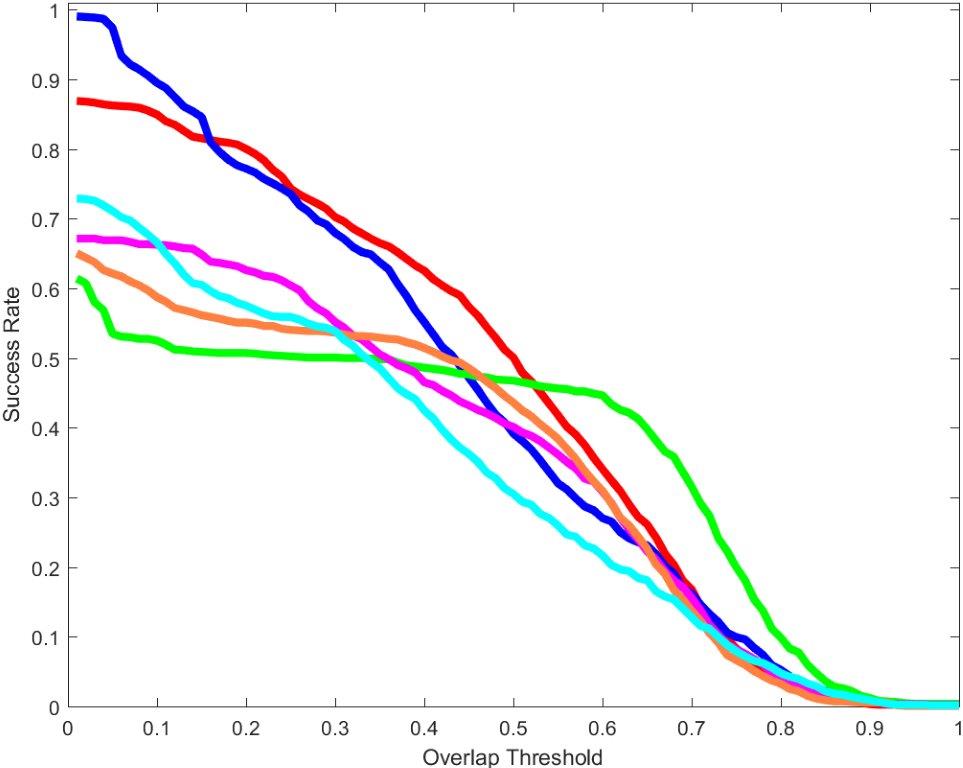}}
\subfigure[FM (QBST, MEEM, STRK)\label{fig:fm}]{\includegraphics[width= 0.24\linewidth]{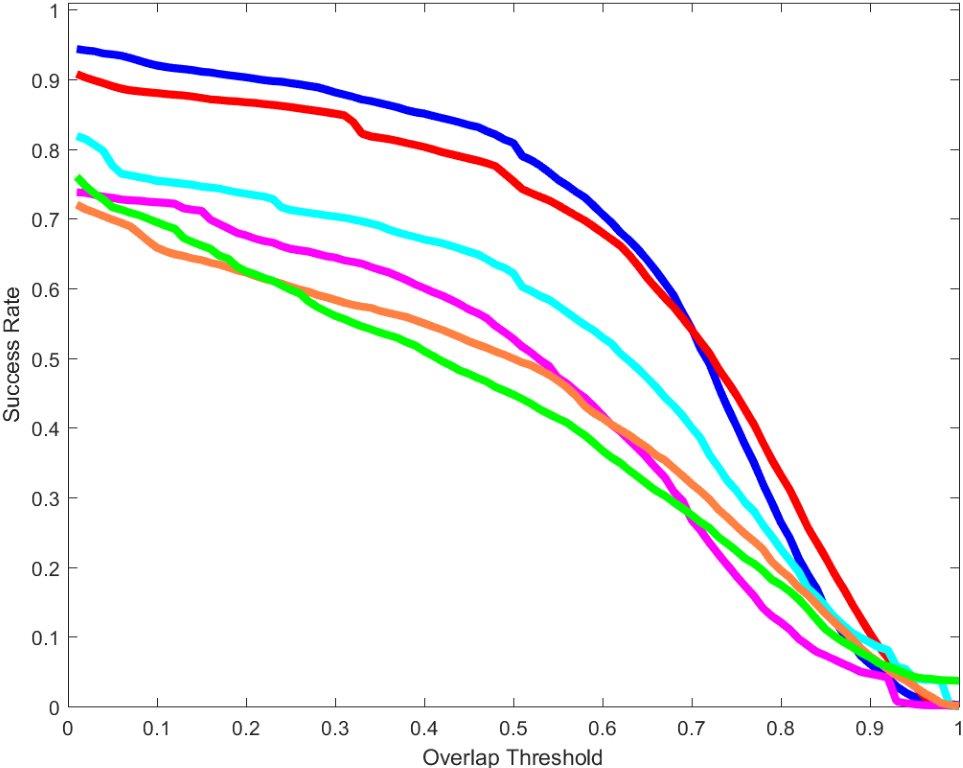}}
\subfigure[MB (MEEM, QBST, STRK)\label{fig:mb}]{\includegraphics[width= 0.24\linewidth]{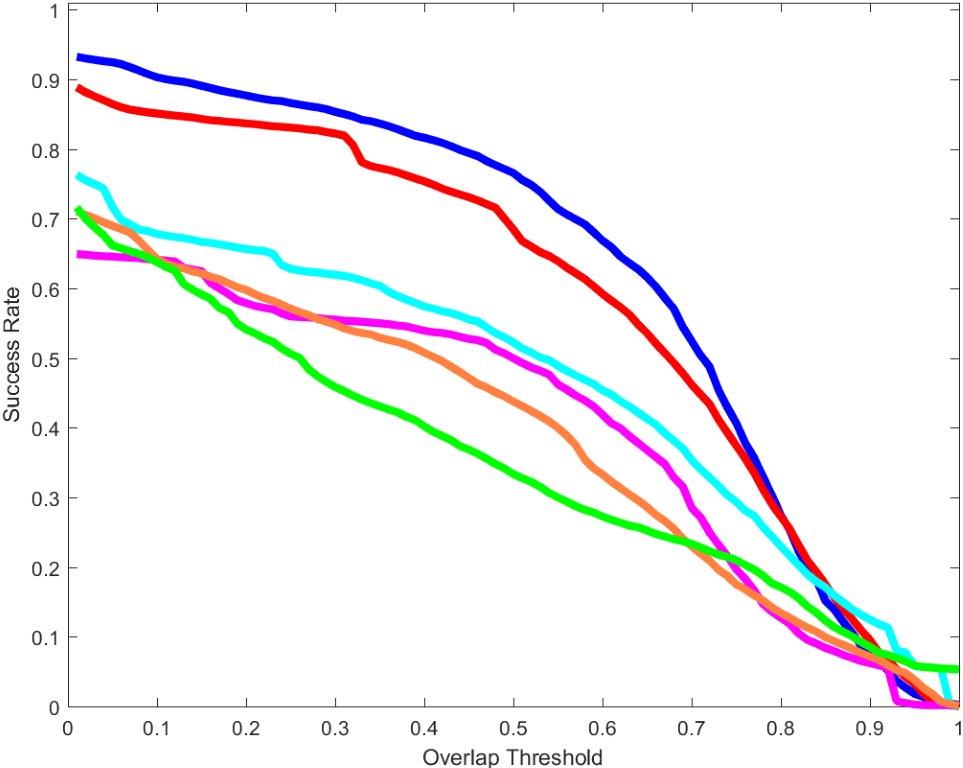}}
\caption{Quantitative evaluation of trackers under different visual tracking challenges (Top three performing trackers are listed in the order of their $AUC$ values). The {\color{red}\textbf{QBST}} is plotted against {\color[rgb]{0,0,1} \textbf{MEEM}}\cite{zhang2014meem}, {\color[rgb]{0,1,1} \textbf{STRK}}\cite{hare2011struck}, {\color[rgb]{0,1,0} \textbf{CXT}}\cite{dinh2011context}, {\color[rgb]{1,0,1} \textbf{TLD}}\cite{kalal2012tracking}, {and \color[rgb]{1,0.5,0.25} \textbf{VTS}}\cite{kwon2011tracking}. QBST outperformed other trackers (except in the MB \ref{fig:mb} category) when dealing with different tracking challenges of OTB-50 \cite{wu2013online} at all of the subcategories. It is shown in \ref{fig:all} that QBST, clearly has a better overall performance compared to other trackers.}
\label{fig:eval_succ_all}
\vspace{-0.5 cm}
\end{figure*}
%%%%%%%%

This section reports on the experiments, first investigating the effect of $\delta$ on the performance of the proposed tracker, and then comparing it with relevant algorithms on benchmark sequences that are commonly used in the literature. The experiments are conducted on 50 challenging video sequences from \cite{wu2013online}, which involves many visual tracking challenges such as illumination variation (IV), scale variation (SV), occlusions (OCC), deformations (DEF), motion blur (MB), fast motion (FM), in-plane rotation (IPR), out-of-play rotation (OPR), out-of-view problem (OV), background clutter (BC) and low resolution (LR). The performance of the tracker is measured by the area under the surface of its success plot ($AUC$), where the success of tracker in time $t$ is determined when the overlap of the tracker target estimation $\mathbf{p}_t$  with the ground truth $\mathbf{p}_t^*$ exceeds a threshold $\tau_{ov}$. Success plot, graphs the success of the tracker against different values of the threshold $\tau_{ov}$ and its $AUC$ is calculated as
\begin{equation}
AUC = \frac{1}{T} \int_0^1 \sum_{t=1}^T \mathds{1} \left( \frac{ | \mathbf{p}_t \cap \mathbf{p}_t^* | }{ | \mathbf{p}_t \cup \mathbf{p}_t^* | } > \tau_{ov} \right) d_{\tau_{ov}},
\end{equation}
where $T$ is the length of sequence, $|.|$ denotes the area of the region, $\cap$ and $\cup$ stands for intersection and union of the regions respectively, and $\mathds{1}(.)$ denotes the step function that returns 1 iff its argument is positive and 0 otherwise. This plot provides an overall performance of the tracker, reflecting target loss, scale mismatches, and localization accuracy. However, to compare the accuracy of compared trackers, precision plot was also presented in Figure \ref{fig:precision}. QBST achieved the average speed of 25.81 fps on a Pentium IV PC @ 3.5 GHz and a Matlab/C++ implementation with no code optimization. HOC and HOG features where used in the implementation and parameters were set at $C = 7, \tau_B = 15, \Delta = 11, n = 1000, m = 130, \tau_l = -0.38 , \tau_u = +0.38$.

First, we investigate the effect of $\delta$ on the ``activeness'' of the tracker, i.e., the amount and quality of information exchange between the oracle and the committee. As discussed earlier, different values of $\delta$ provide different trackers in a wide spectrum of a single classifier to an ensemble, covering various degrees of data exchange. Figure \ref{fig:delta} compared several values of the parameter $\delta$ to illustrate this effect. The optimal value $\delta = 0.38$ is obtained by 5-fold cross-validation on the current dataset. Also, to demonstrate the effectiveness of data exchange scheme, we replaced it with a random query scheme (called \textit{random} in Fig. \ref{fig:delta}), that query each sample from either the oracle or the committee based on a Bernoulli distribution with $p = 0.5$. It is evident from the figure, that \textit{random} and \textit{oracle-only} ($\delta = 1$) versions of QBST are not sophisticated enough to handle various tracking challenges. On the other hand, when the $\delta$ mostly rely on oracle to fix the committee results ($\delta = 0.8$), the performance of the tracker is not as good as the \textit{oracle-only} case ($\delta = 1$), due to large size of the $\mathcal{U}_t$ that doesn't provide a concrete direction for committee to update itself. In the case of $\delta = 1$, the committee is updated with all of the available labels.

To establish a fair comparison with the state-of-the-art, we select some of the most popular discriminative and generative trackers (according to a recent large benchmark \cite{wu2013online} and the literature): \textit{(i)} CXT \cite{dinh2011context} that utilizes context to better localize the target, \textit{(ii)} MEEM \cite{zhang2014meem} that can role-back bad updates by selecting among the snapshots of the classifier throughout the tracking, \textit{(iii)} STRUCK \cite{hare2011struck} that uses an structured SVM to marge sampling and labeling to avoid the label-noise problem, \textit{(iv)} TLD \cite{kalal2012tracking} that use auxiliary classifiers to monitor false-positives and false-negatives of the labeler, and \textit{(iv)} VTS \cite{kwon2011tracking} that sample from a pool of appearance and motion models to construct the most suitable tracker for the current observation.

We perform a benchmark on the whole videos of the dataset (Fig. \ref{fig:precision}), along with partial subsets of the dataset with a distinguishing attribute (Fig. \ref{fig:eval_succ_all}) to evaluate the tracker performance under different situations. 

%%%%%%%%
\begin{figure*}[!t]
\centering
\subfigure[Tracking results of sequence \textit{FaceOcc2} and \textit{Walking2} with severe occlusions]
{\includegraphics[width= 0.16\linewidth]{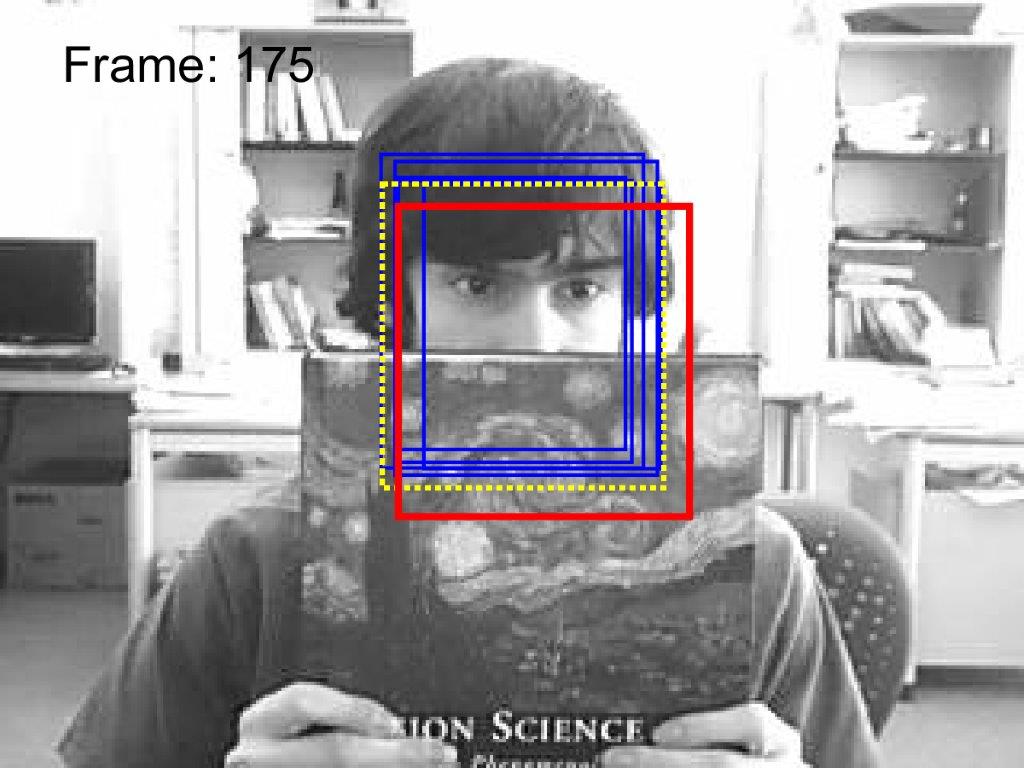}
\includegraphics[width= 0.16\linewidth]{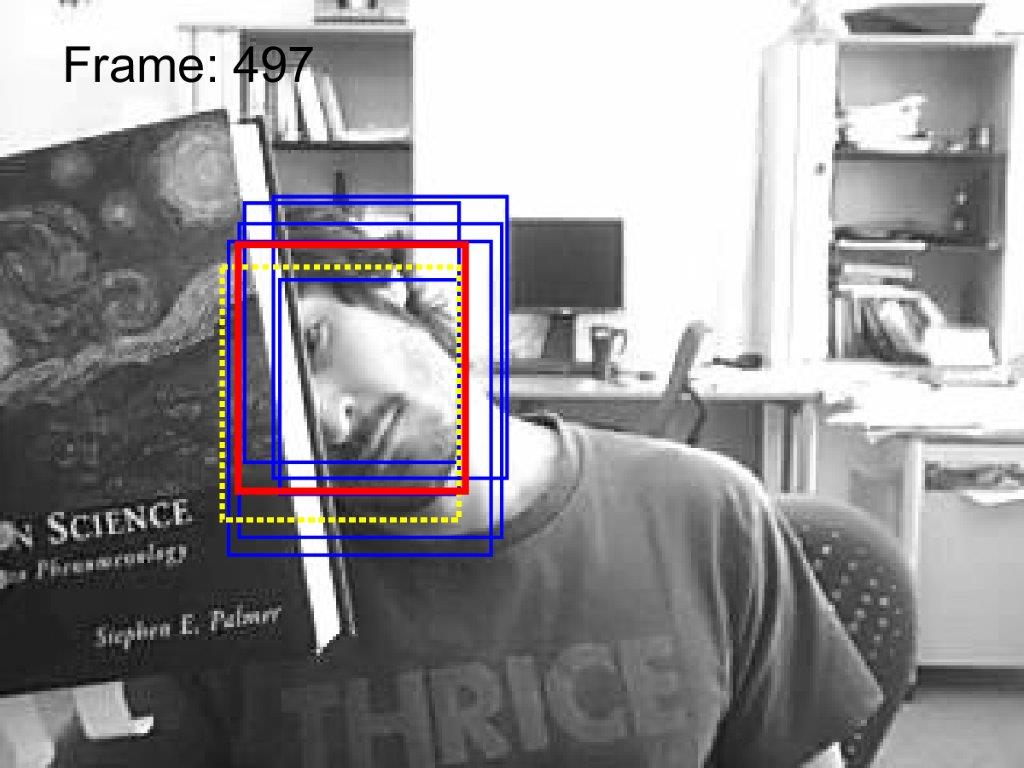}
\includegraphics[width= 0.16\linewidth]{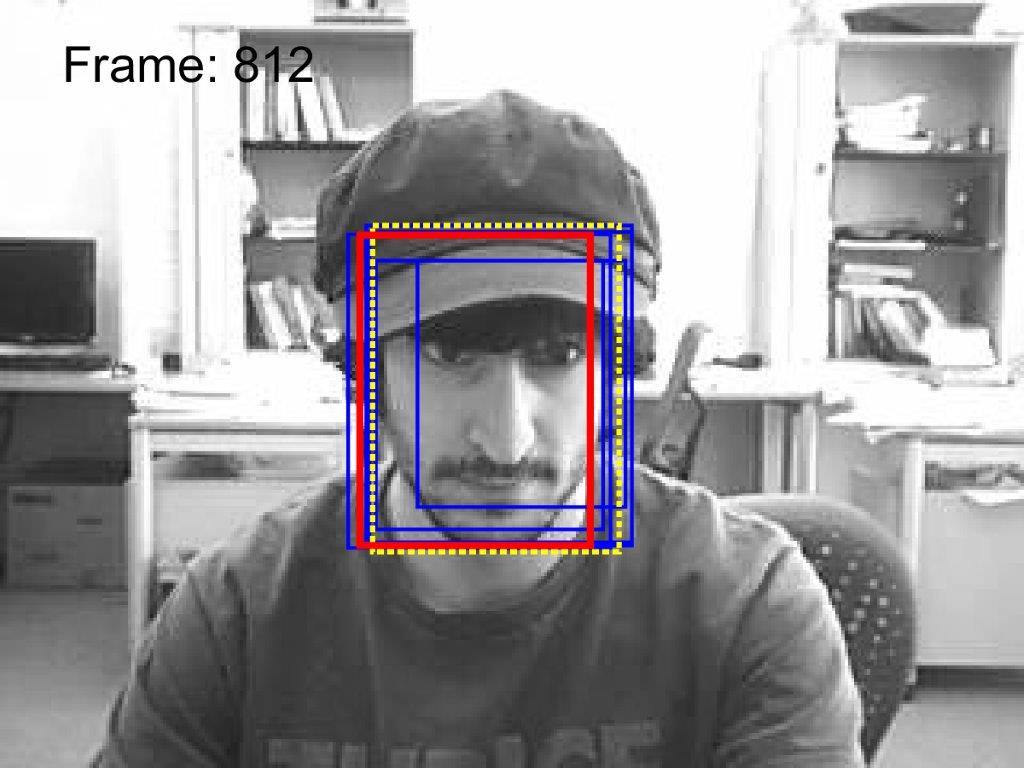}
\includegraphics[width= 0.16\linewidth]{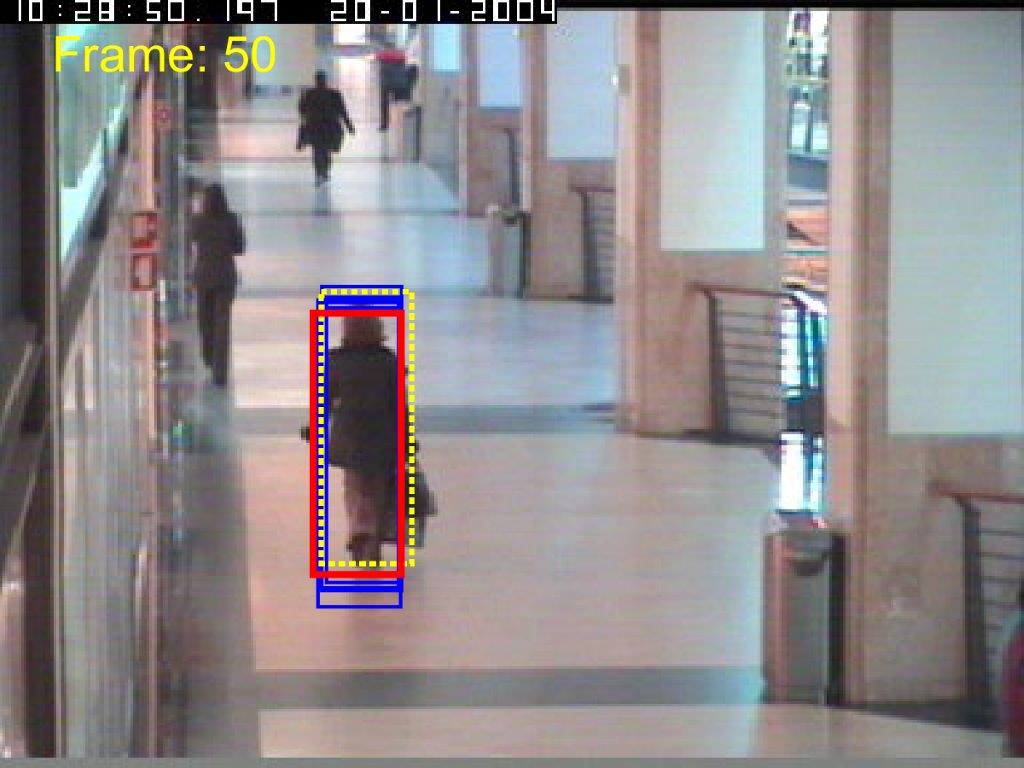}
\includegraphics[width= 0.16\linewidth]{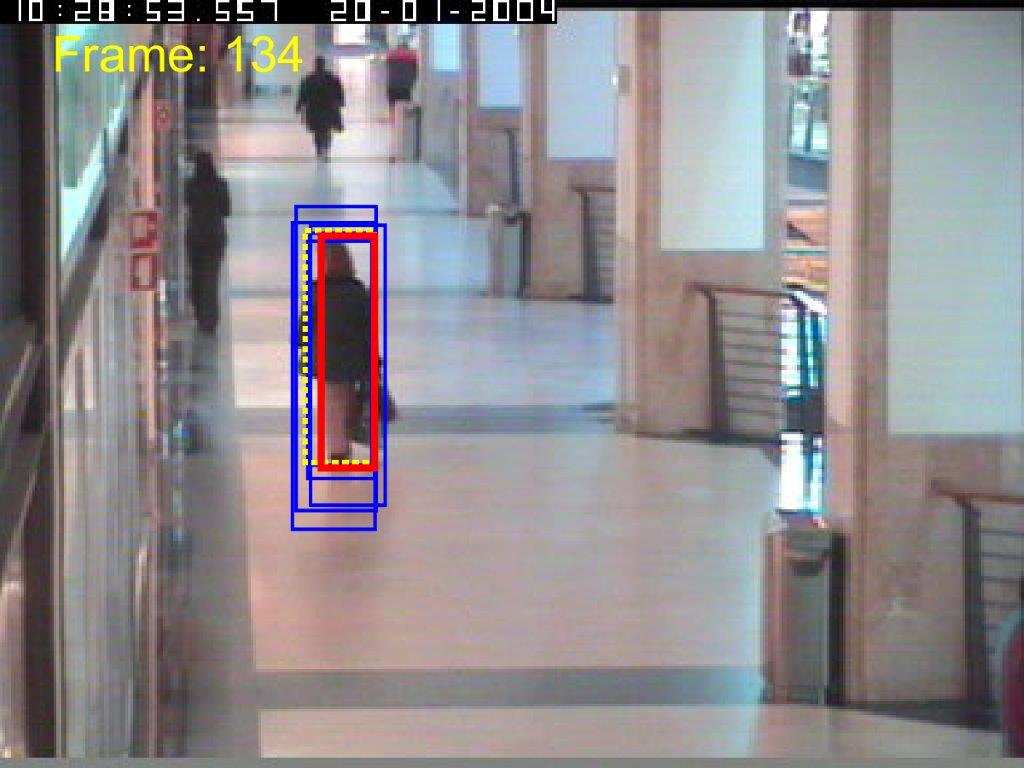}
\includegraphics[width= 0.16\linewidth]{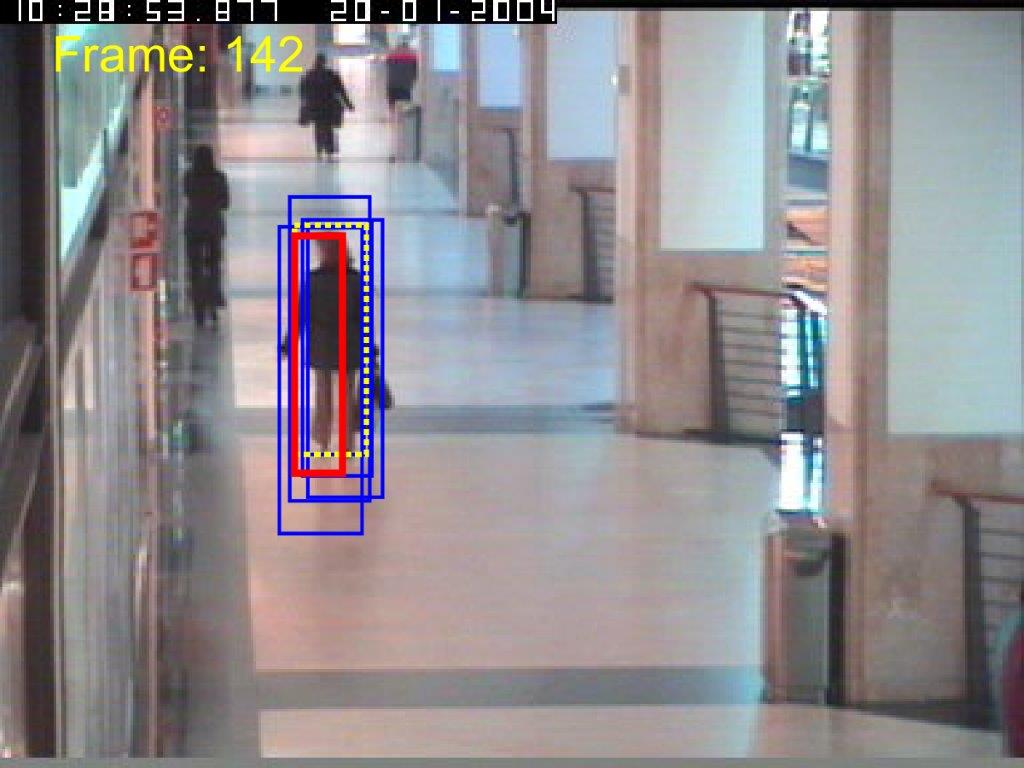}
}
\subfigure[Tracking results of sequence \textit{Basketball} and \textit{Skating1} with deformations]
{\includegraphics[width= 0.16\linewidth]{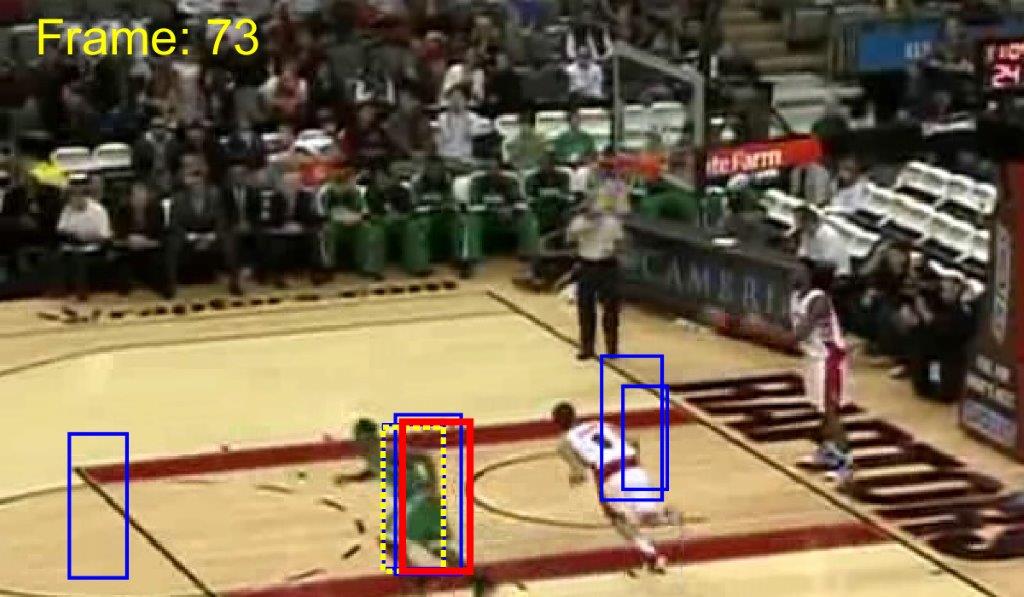}
\includegraphics[width= 0.16\linewidth]{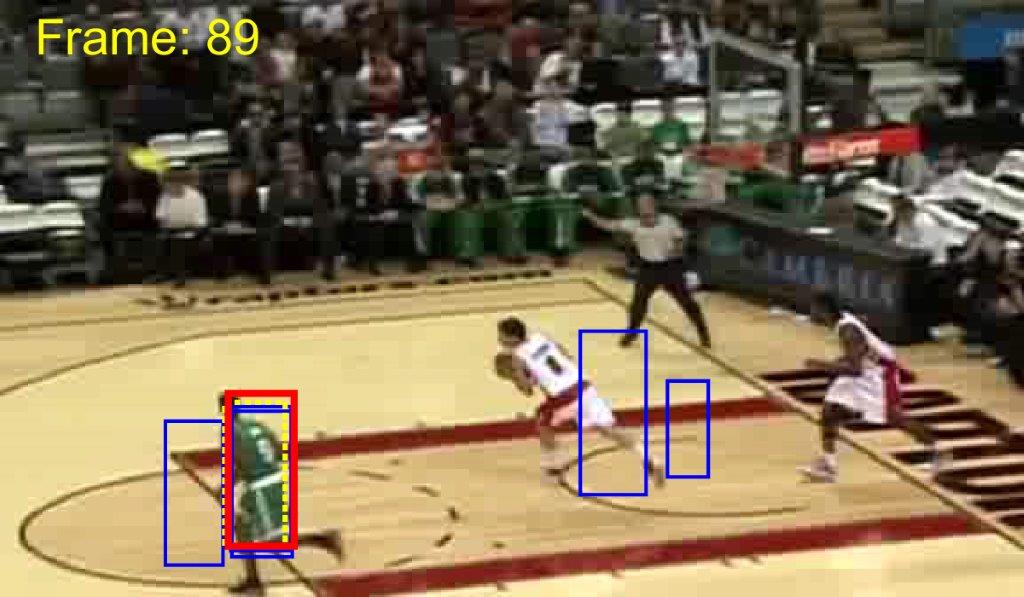}
\includegraphics[width= 0.16\linewidth]{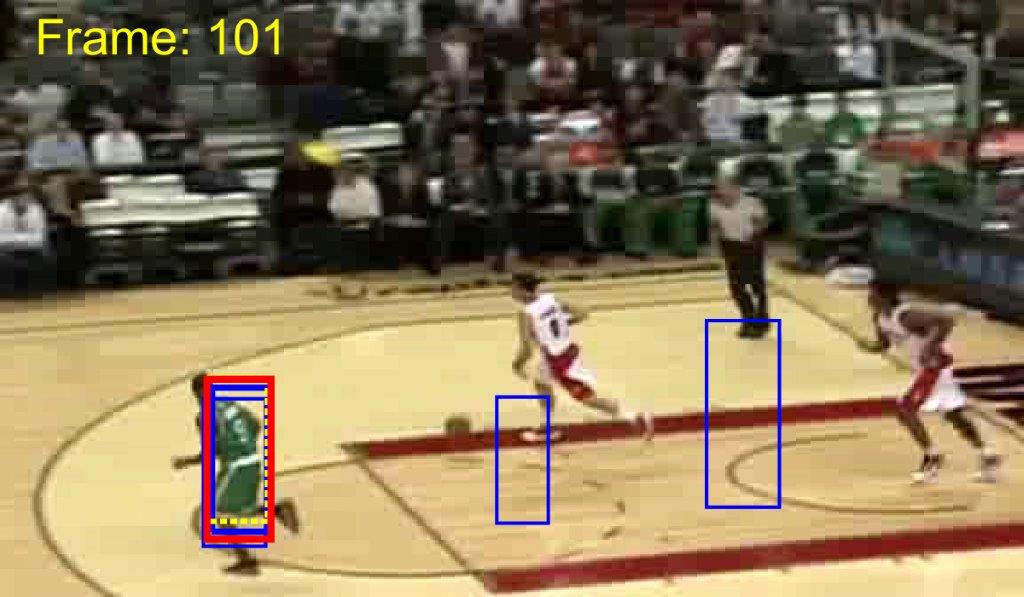}
\includegraphics[width= 0.16\linewidth]{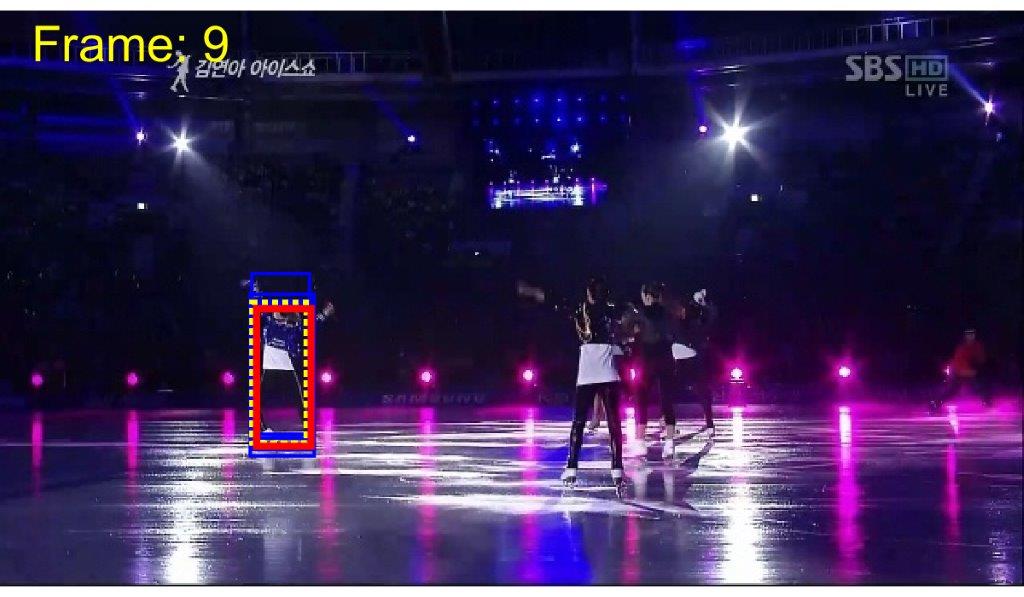}
\includegraphics[width= 0.16\linewidth]{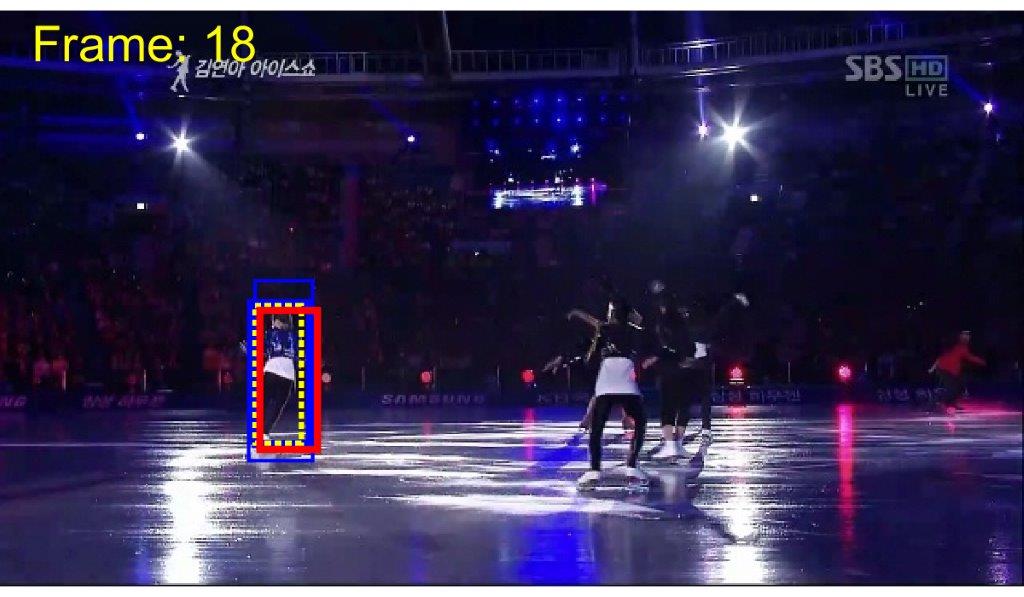}
\includegraphics[width= 0.16\linewidth]{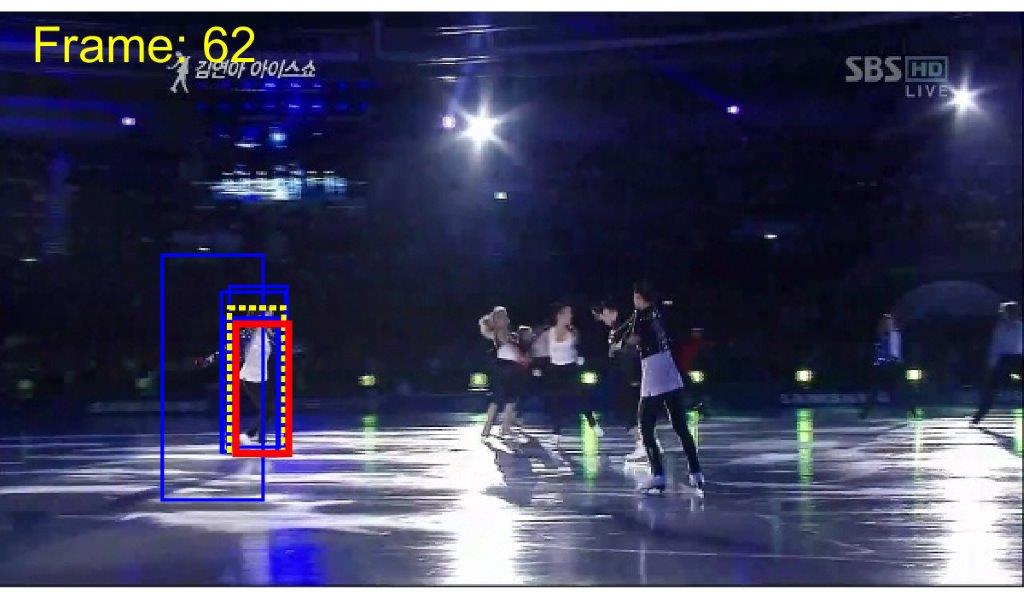}
}
\subfigure[Tracking results of sequence \textit{Girl} and \textit{Ironman} with in-plane and out-of-plane rotations]
{\includegraphics[width= 0.16\linewidth]{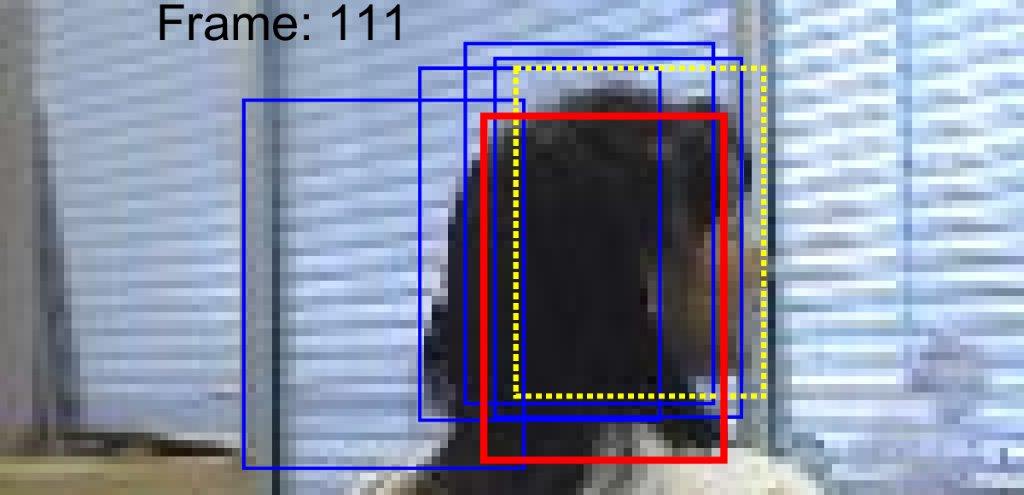}
\includegraphics[width= 0.16\linewidth]{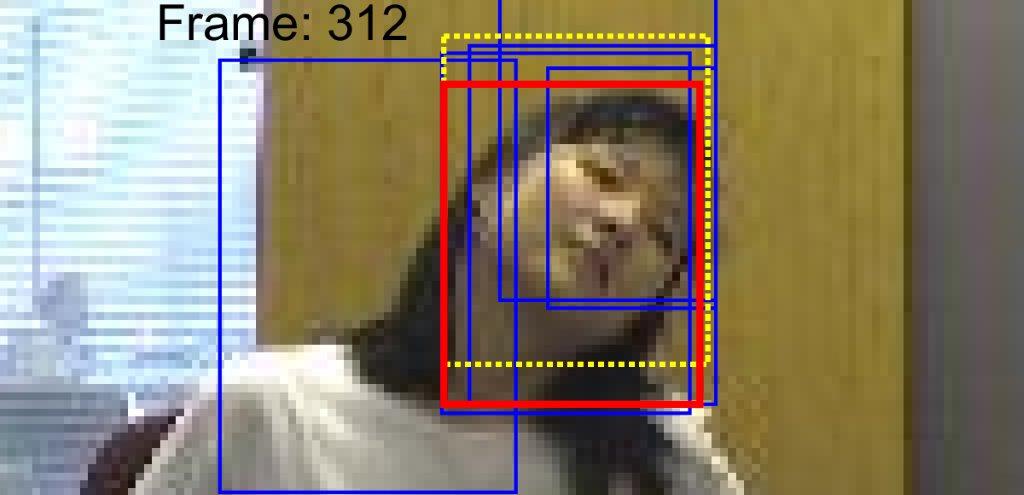}
\includegraphics[width= 0.16\linewidth]{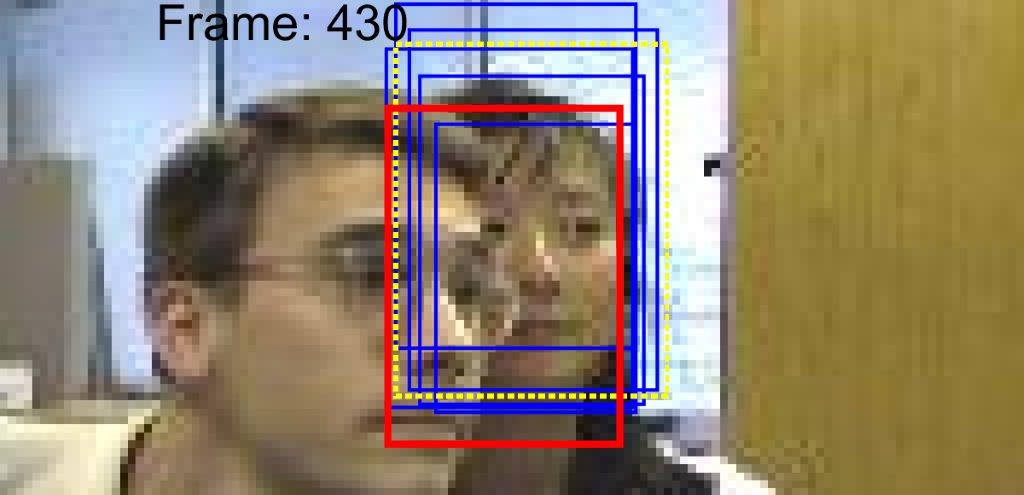}
\includegraphics[width= 0.16\linewidth]{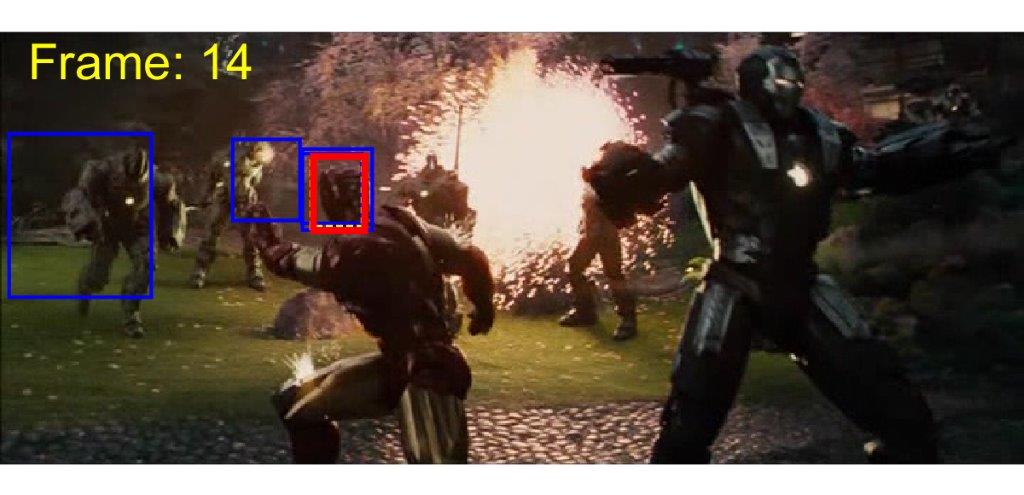}
\includegraphics[width= 0.16\linewidth]{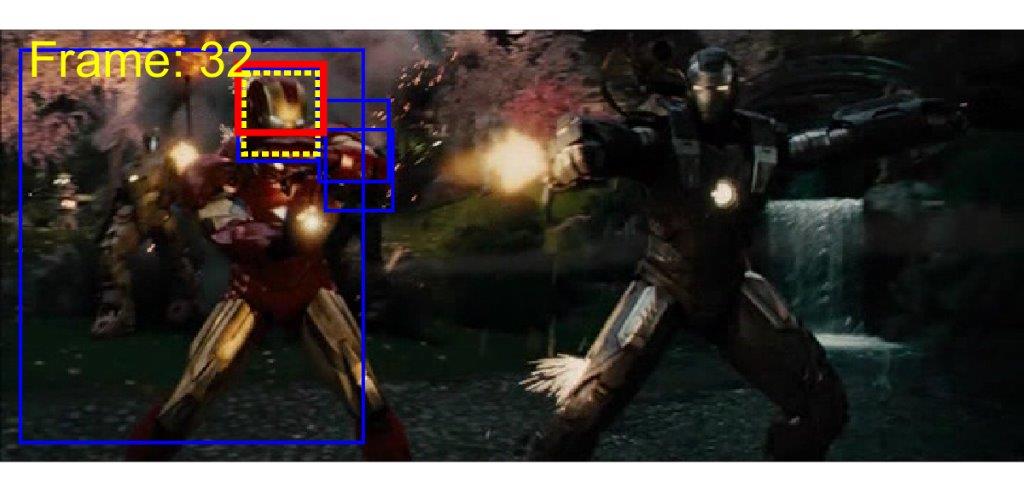}
\includegraphics[width= 0.16\linewidth]{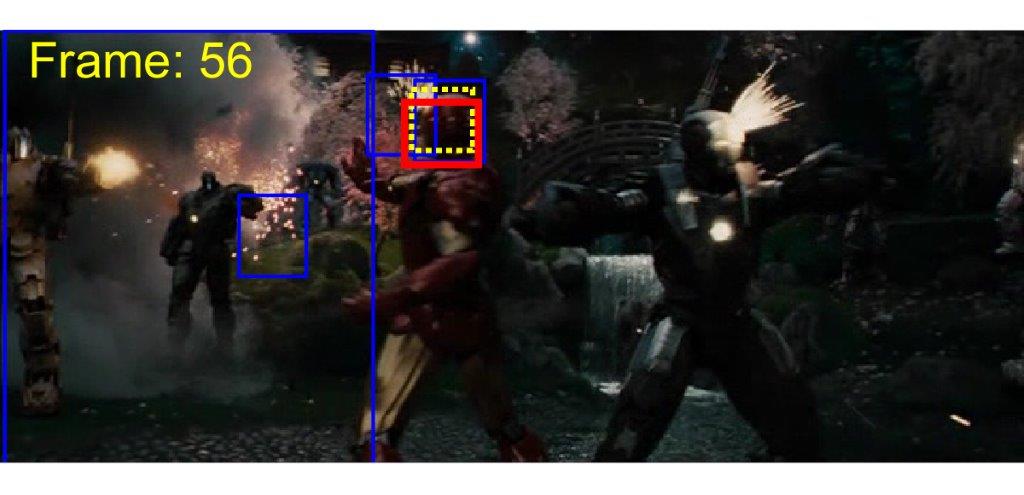}
}
\subfigure[Tracking results of sequence \textit{Singer2}, \textit{Shaking} and \textit{CarDark} with drastic illumination changes]
{\includegraphics[width= 0.16\linewidth]{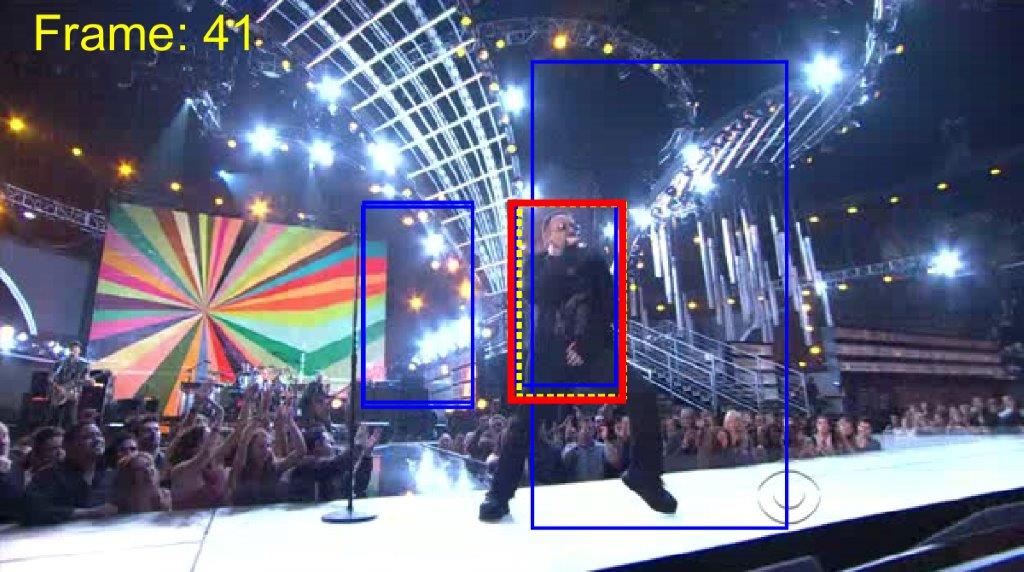}
\includegraphics[width= 0.16\linewidth]{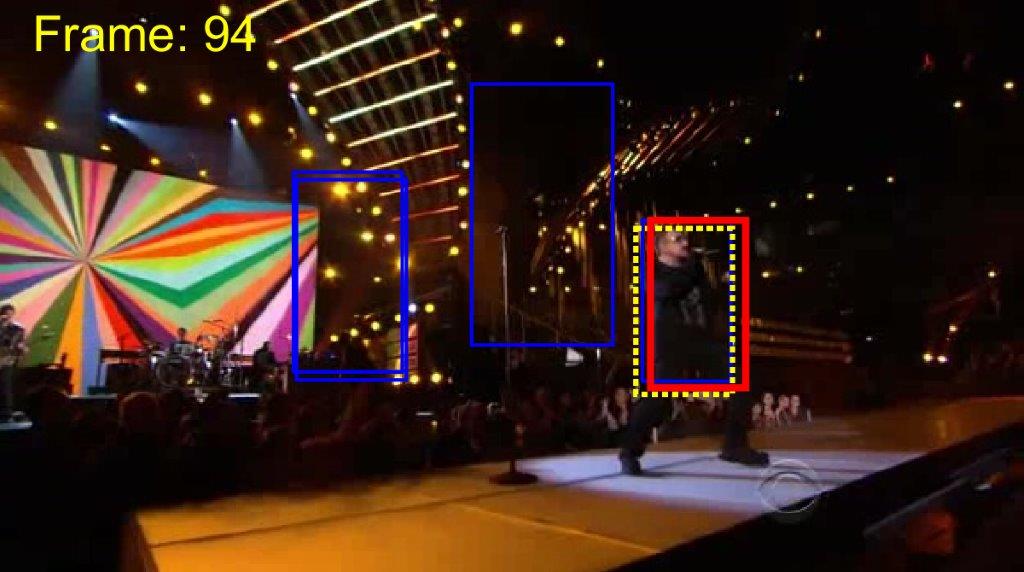}
\includegraphics[width= 0.16\linewidth]{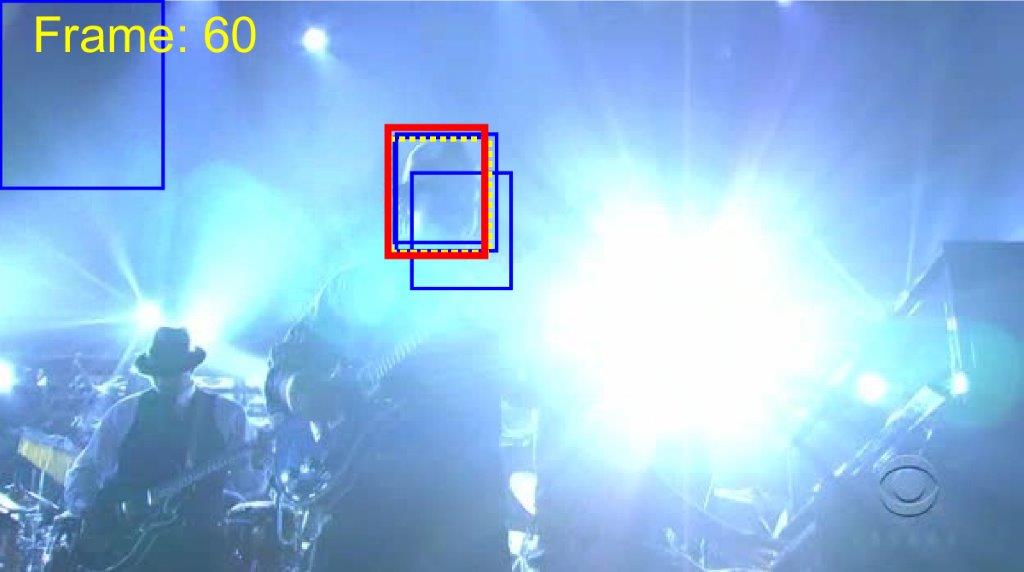}
\includegraphics[width= 0.16\linewidth]{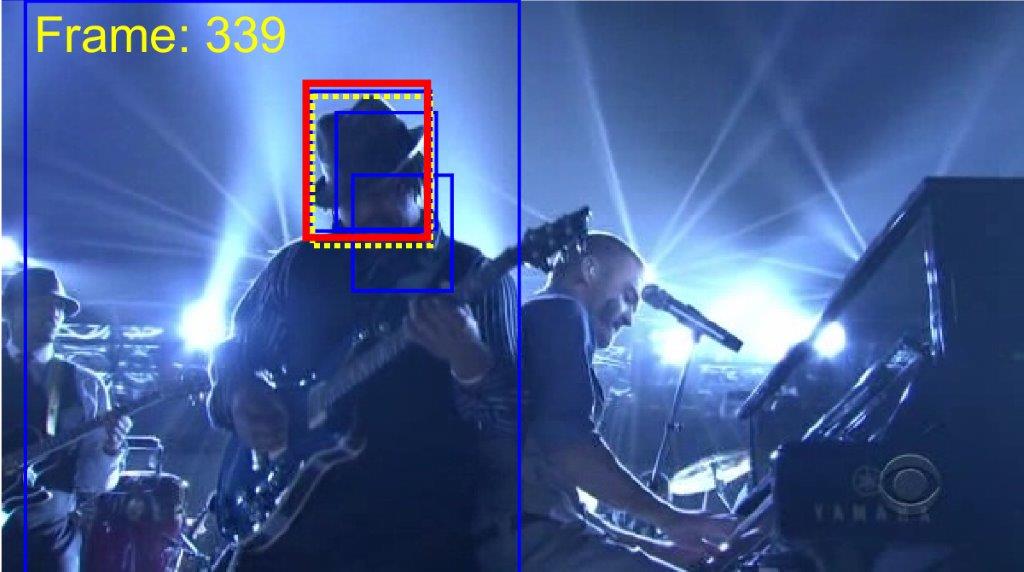}
\includegraphics[width= 0.16\linewidth]{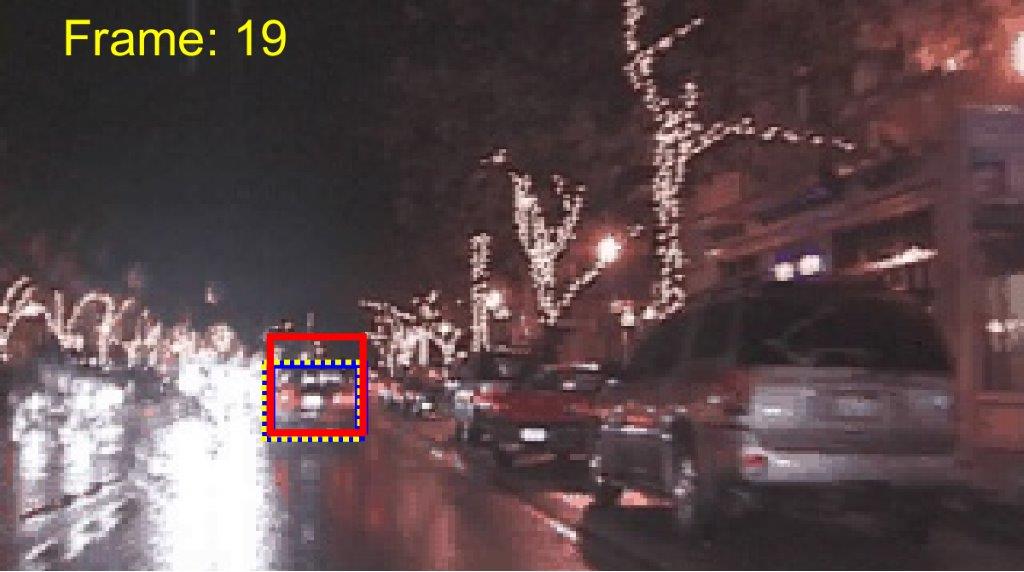}
\includegraphics[width= 0.16\linewidth]{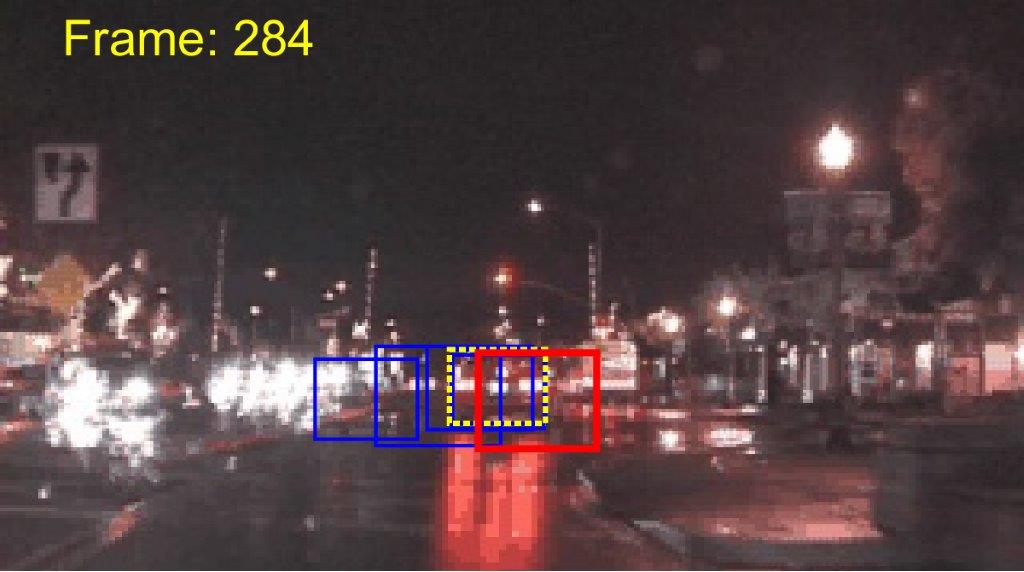}
}
\subfigure[Tracking results of sequence \textit{David3} with background clutter]
{\includegraphics[width= 0.16\linewidth]{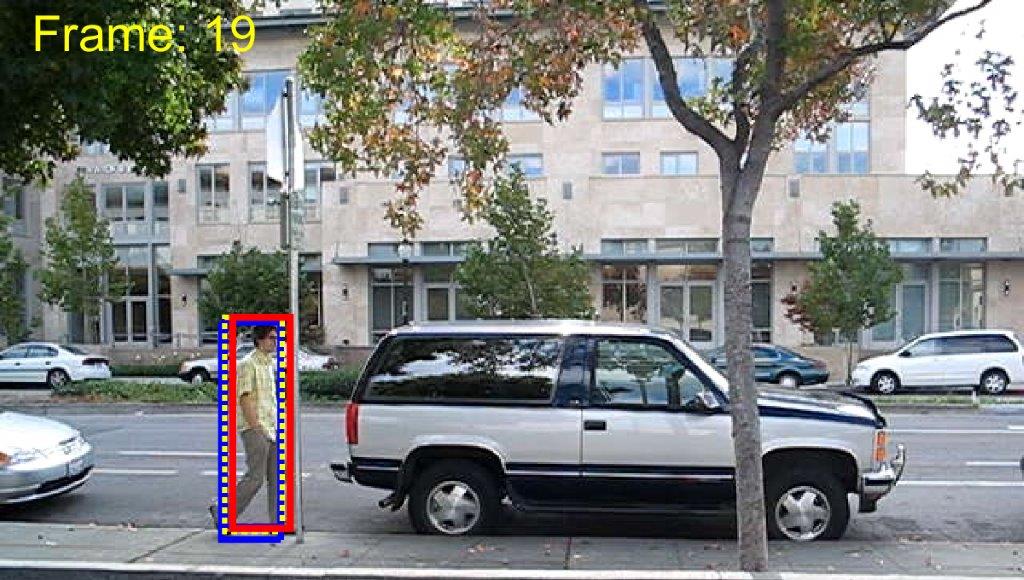}
\includegraphics[width= 0.16\linewidth]{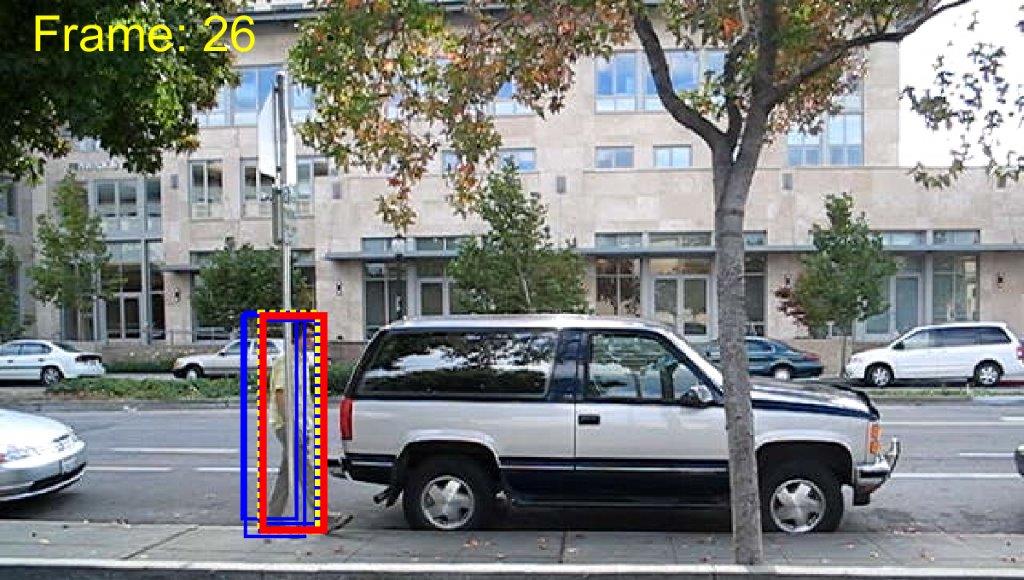}
\includegraphics[width= 0.16\linewidth]{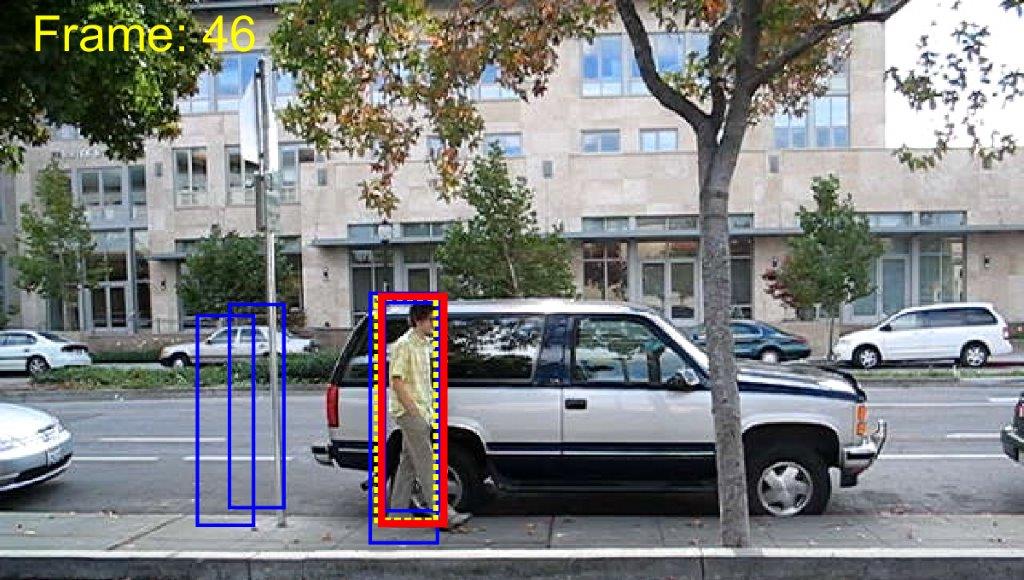}
\includegraphics[width= 0.16\linewidth]{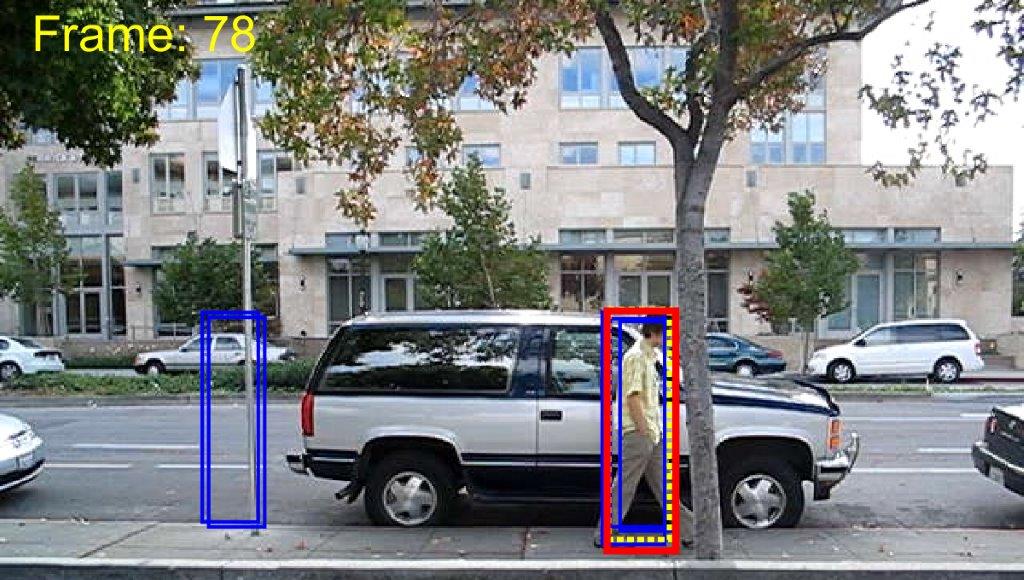}
\includegraphics[width= 0.16\linewidth]{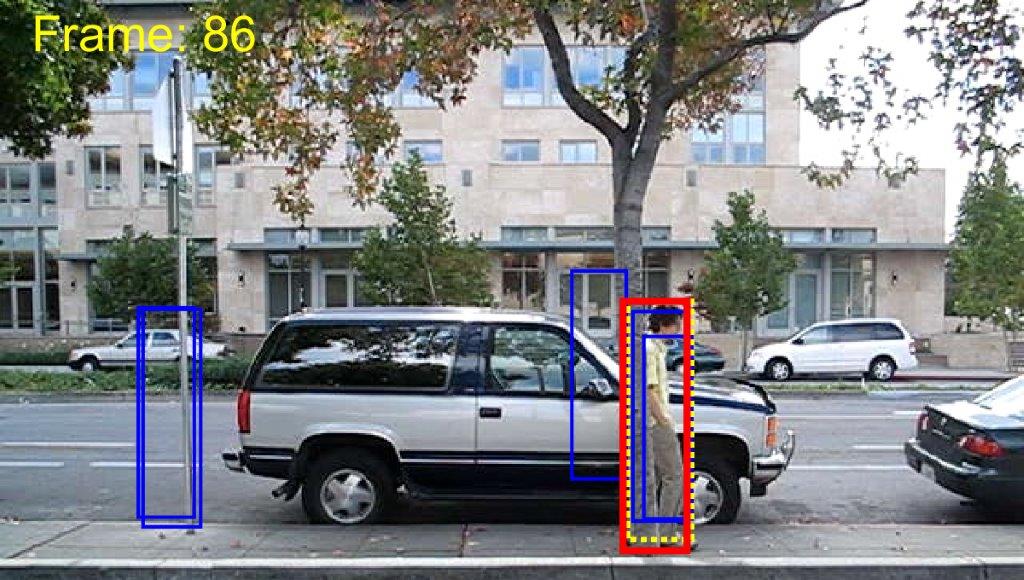}
\includegraphics[width= 0.16\linewidth]{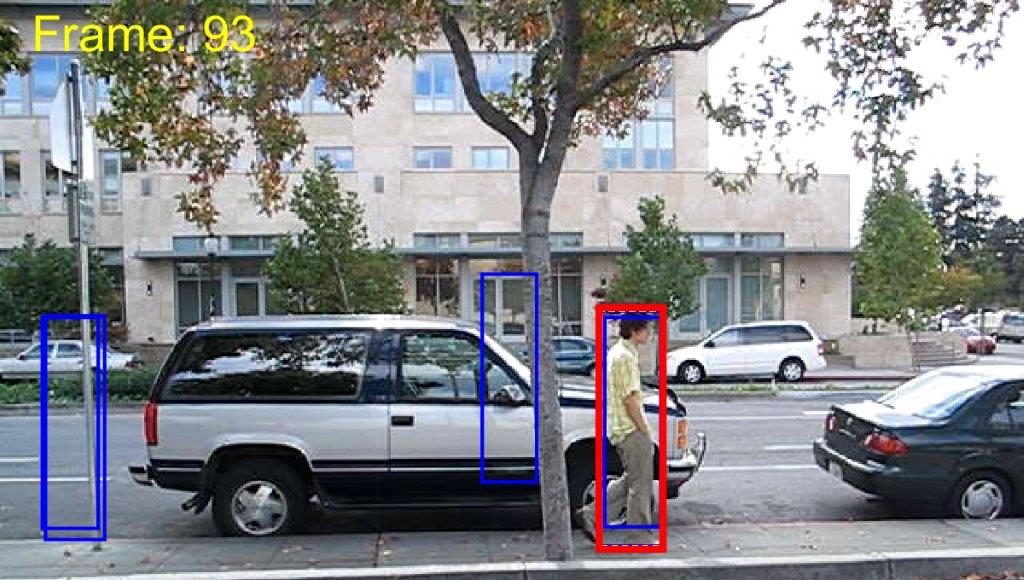}
}
\caption{Sample tracking results of evaluated algorithms on several challenging video sequences. In these sequences the red box depicts the QBST against other trackers (blue). The ground truth is illustrated with yellow dashed box. The results are available in the \texttt{webpage}.
}
\label{fig:eval_qual}
\vspace{-0.5 cm}
\end{figure*}
%%%%%%%%

Figure \ref{fig:eval_succ_all} presents the success plot of QBST along with other state-of-the-art trackers for all sequences, and its subcategories, each focusing on a certain challenge of the visual tracking. As demonstrated in \ref{fig:all}, QBST has the best overall performance among investigated trackers on this dataset. While this algorithm has a clear edge in handling many challenges, its performance is comparable with MEEM \cite{zhang2014meem} in the case of background clutter (Fig. \ref{fig:bc}), low resolution (Fig. \ref{fig:lr}), and out-of-view (Fig \ref{fig:ov}). It is also evident that both QBST and MEEM are troubled with fast motion and motion blur (Figs. \ref{fig:fm} and \ref{fig:mb}), since neither of them has an explicit motion model to handle these cases, however, MEEM handled motion blur better than the proposed QBST. Interestingly, it is observed that in most of the subcategories that QBST is clearly better than the other trackers, the success plot of QBST starts with a plateau and later has a sharp drop around $\tau_{ov} = 0.6$. This means that QBST provides high-quality localization (i.e., bigger overlaps with the ground truth). Similarly, it is evident that the proposed algorithm shows a graceful degradation in low-resolution scenario (\ref{fig:lr}), and although it does not provide a high-quality localization for smaller/low-resolution targets, it is able to keep tracking them. Moreover, the starting values (i.e., when $\tau_{ov} \rightarrow 0$) are in most cases higher than other trackers, indicating that the number of frames where the target is lost is lower than other trackers. Another interesting observation is that QBST and MEEM, both based on an ensemble of self-adjusting classifiers, outperform other trackers in most of the tracking challenges. This finding highlights the importance of further research on the ensemble-based trackers.

These results are supported by the precision plot depicted in Figure \ref{fig:precision}. The precision plot compares the number of frames that a tracker has certain pixels of displacement. It is shown in this plot that QBST usually keeps the localization error under 10 pixels. A qualitative comparison of QBST versus other trackers is presented in Figure \ref{fig:eval_qual}.

The high accuracy of the QBST (Fig. \ref{fig:precision}) can be attributed to using an ensemble of classifiers, and low label-noise obtained by the devised scheme to label the uncertain samples. The effective coverage of the version space, improves the diversity of committee, which in turn boosts the accuracy of the obtained strong classifier. Additionally, the weighted vote of classifiers, robustify the tracking in the already-explored regions of the version space, as it is evident in sequences with background clutter (Fig. \ref{fig:bc}). One the other hand, actively balancing long-term memory of the oracle with the short-term memory of the committee, empowers the tracker to handle occlusions (Fig. \ref{fig:occ}), deformations (Fig. \ref{fig:def}), and illumination variations (Fig. \ref{fig:iv}) well. It should be noted that not defining an effective region-of-interest (e.g., using motion models) was due to the fact that we wish to investigate the role of boosting in sampling the version space, however, this decision affect the tracker performance of the tracker in sequences with abrupt/rapid motions (Fig. \ref{fig:fm} and (Fig. \ref{fig:mb})). Investigating the interaction of QBST engine with motion models and possible improvements obtained from that, are one of the future directions that this ongoing project is seeking to pursue.

\section{Conclusion}
\label{sec:conclusion}
In this study, we proposed query-by-boosting tracker that maintains a diverse committee of classifiers to the label of the samples and queries the most disputed labels --which are the most informative ones-- from a long-term memory oracle. By using the query-by-boosting principles, this tracker is efficiently shrinking the version space using the set of models in the committee. In addition by using boosting in updating the committee and adjusting their weight in the labeling process, the label noise problem is decreased because of: 
\textit{(i)} the local search in the version space to improve the quality of the models, 
\textit{(ii)} updating classifiers only with the most informative samples (obtained by active learning) and most efficient samples (obtained by boosting), 
and \textit{(iii)} the weighted voting to assign the label to each sample based on the confidence of different committee members on their decision. 
By using the weighted votes of the committee, in turn, the problem of equal weights for the samples are addressed, and a good approximation of the target location is acquired even without dense sampling.  
The active learning scheme also manages the balance between short-term and long-term memory by recalling the label from long-term memory when the short-term memory is not clear about the label (due to forgetting the label or insufficient data). This also reduces the dependence of the tracker on a single classifier while breaking the self-learning loop.  

The result of the experiment on a recent large benchmark \cite{wu2013online} demonstrates the superior tracking performance of the proposed tracker compared with the state-of-the-art generative trackers and tracking-by-detection algorithms. In the next step, we will investigate the interaction of the proposed tracker with motion models or other sampling strategies.   

\section*{Acknowledgment}
This article is based on results obtained from a project commissioned by the New Energy and Industrial Technology Development Organization (NEDO).

\bibliographystyle{IEEEtran}
\bibliographystyle{model2-names}
\def\IEEEbibitemsep{3.9pt plus 1pt}
\bibliography{qbstbib}

\end{document}